\documentclass[lettersize,journal]{IEEEtran}
\usepackage{amsmath,amsfonts}
\usepackage{algorithmic}
\usepackage{algorithm}
\usepackage{array}
\usepackage[caption=false,font=normalsize,labelfont=sf,textfont=sf]{subfig}
\usepackage{textcomp}
\usepackage{stfloats}
\usepackage{url}
\usepackage{verbatim}
\usepackage{graphicx}
\usepackage{cite}
\usepackage{booktabs} 
\usepackage{hyperref} % 添加此行以启用超链接

\usepackage{multirow}
\usepackage[table]{xcolor}
\usepackage{caption}
\usepackage{orcidlink}

\usepackage{xcolor} 
\begin{document}

\title{DRRNet: Macro–Micro Feature Fusion and Dual Reverse Refinement for Camouflaged Object Detection}

\author{%
	Jianlin Sun\orcidlink{0009-0008-3067-6278}, %
	Xiaolin Fang\orcidlink{0000-0002-0164-2596},~\IEEEmembership{Member,~IEEE}, 
	Juwei Guan\orcidlink{0000-0001-6337-1253}, %
	Dongdong Gui\orcidlink{0009-0007-8715-2398}, %
	Teqi Wang\orcidlink{0000-0001-6273-8105}, %
	Tongxin Zhu\orcidlink{0000-0002-6664-5333},~\IEEEmembership{Member,~IEEE}, 
	\thanks{
		 Jianlin Sun, Xiaolin Fang, Juwei Guan, Dongdong Gui, Teqi Wang and Tongxin Zhu are with the School of Computer Science and Engineering, Southeast University, Nanjing 211189, China, and also with the Key Laboratory of Computer Network and Information Integration (Southeast University), Ministry of Education, China (e-mail: jianlinsun@seu.edu.cn; xiaolin@seu.edu.cn; jvguan@seu.edu.cn, dd.gui@seu.edu.cn, wangteqi@seu.edu.cn, zhutongxin@seu.edu.cn).
	}
	\thanks{
		Corresponding author: Xiaolin Fang (xiaolin@seu.edu.cn).
	}
}

\maketitle

\begin{abstract}
The core challenge in Camouflage Object Detection (COD) lies in the indistinguishable similarity between targets and backgrounds in terms of color, texture, and shape. This causes existing methods to either lose edge details (such as hair-like fine structures) due to over-reliance on global semantic information or be disturbed by similar backgrounds (such as vegetation patterns) when relying solely on local features. We propose DRRNet, a four-stage architecture characterized by a ``context-detail-fusion-refinement'' pipeline to address these issues. Specifically, we introduce an Omni-Context Feature Extraction Module to capture global camouflage patterns and a Local Detail Extraction Module to supplement microstructural information for the full-scene context module. We then design a module for forming dual representations of scene understanding and structural awareness, which fuses panoramic features and local features across various scales. In the decoder, we also introduce a reverse refinement module that leverages spatial edge priors and frequency-domain noise suppression to perform a two-stage inverse refinement of the output. By applying two successive rounds of inverse refinement, the model effectively suppresses background interference and enhances the continuity of object boundaries. Experimental results demonstrate that DRRNet significantly outperforms state-of-the-art methods on benchmark datasets. Our code is available at \url{https://github.com/jerrySunning/DRRNet}.
\end{abstract}

\begin{IEEEkeywords}
Camouflage object detection, Hierarchical feature fusion, Edge-aware segmentation, Frequency-domain calibration, Dual Reverse Refinement.
\end{IEEEkeywords}

\section{Introduction}

Camouflaged Object Detection (COD) is a highly challenging task in the field of computer vision, whose primary goal is to accurately identify targets in complex scenes that are highly similar to their backgrounds in terms of color, texture, or shape\cite{ref1,ref2}. Such targets are widely encountered in domains such as military reconnaissance, ecological conservation, and industrial quality inspection—for example, in military camouflage equipment\cite{ref3}, concealed species in wildlife protection, and defect detection in industrial parts\cite{ref4,ref5,ref6,ref7}. Traditional camouflaged object detection methods rely on handcrafted low-level features to capture subtle differences in texture and color\cite{ref8,ref9,ref10}. However, due to the extremely high similarity between the camouflaged target and its background, these methods often fail to adequately extract key features, leading to a significant degradation in detection performance.

\label{sec:intro}
\begin{figure}[htbp]
	\centering
	\includegraphics[width=1\linewidth]{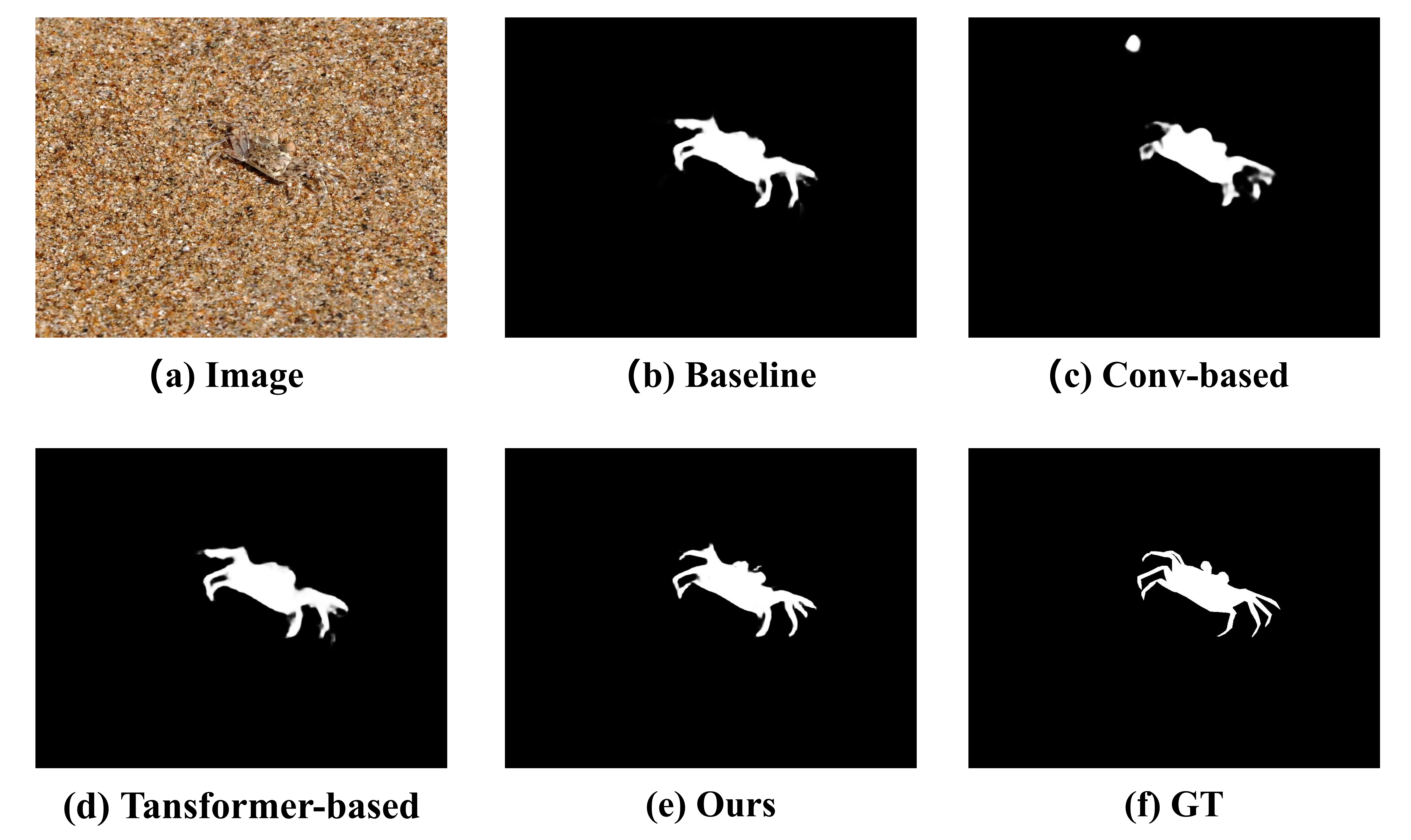}
	\caption{Visual comparison of models utilizing various types of information. (b) Baseline model. (c) Convolution-based model, which fails to accurately detect regions where the crab's legs closely blend with the environment. (d) Transformer-based model, which yields acceptable results, yet remains deficient in local detail. (e) Our model, employing a four-stage framework of panoramic perception – detail mining – cross-level fusion – dual reverse calibration, produces the best results.}
	\label{fig:intro}
	
\end{figure}

Deep learning has propelled advances in COD via the hierarchical feature–extraction capabilities of CNNs (e.g., SINet’s\cite{ref11} dilated‐convolution Search Module, C2F-Net’s\cite{ref12} cross‐level fusion, and FAPNet’s\cite{ref13} multi‐scale aggregation). However, the inherently limited receptive fields of convolutional kernels hinder the modeling of long‐range dependencies\cite{ref14,ref15}, resulting in blurred object boundaries. Conversely, Transformers—despite their strength in capturing global context through self‐attention—often sacrifice local detail when processing images as patches, rendering them insensitive to small‐scale or weak‐edge targets.

To balance global semantics and local details, recent works have sought to integrate the strengths of CNNs and Transformers\cite{ref19,ref20,ref21,ref22}. For example, the CSWin Transformer\cite{ref17} employs cross-shaped window attention to capture both global context and local structures, yet its decoder relies solely on simple skip connections, leading to suboptimal feature integration. FSEL\cite{ref18} proposes a frequency–space entanglement learning framework that fuses global frequency-domain features with local spatial features via entangled Transformer blocks; however, its decoder still follows a traditional U-Net\cite{ref16} design and fails to recover fine-grained contours in complex scenes. We note that the core issue in these approaches is an imbalance between encoding and decoding: encoder architectures have become increasingly sophisticated, while decoder designs remain overly simplistic (e.g., bilinear interpolation), preventing full feature reconstruction.

Motivated by the above observations, we introduce DRRNet, a unified framework that seamlessly integrates multi-scale contextual perception, micro-detail mining, and dual reverse refinement decoding. Concretely, our encoder first routes multi-scale backbone features into an OmniContext Module and a MicroDetail Module to extract complementary global and local cues. These cues are then jointly modeled and fused in both spatial and frequency domains by the Macro–Micro Fusion module. The deepest fused features are decoded by a Global Rough Decoder to produce an initial coarse prediction map. Finally, a Dual Reverse Refinement Module performs two successive rounds of spatial edge prior calibration and frequency-domain noise suppression, iteratively attenuating background interference and reinforcing boundary continuity for precise camouflaged object segmentation.

Extensive experiments on three COD benchmark datasets demonstrate that, compared to dozens of state-of-the-art COD methods, our DRRNet exhibits clear advantages across multiple evaluation metrics. Our main contributions are summarized as follows:
\begin{itemize}
	\item To achieve the collaborative representation of panoramic context and microscopic details, we designed the OCM and MDM modules. The OCM module effectively captures multi-scale context through a multi-branch mechanism, while the MDM module mines microscopic detail features to enhance sensitivity to subtle differences.
\end{itemize}
\begin{itemize}
	\item To better fuse global and local features, we implemented the MMF module, which performs joint spatial and frequency domain modeling and refines fusion via convolution dimensionality reduction and gated convolution.
\end{itemize}
\begin{itemize}
	\item We introduced the GRD and DRRM modules as decoders. The GRD module quickly decodes global image information, and the DRRM module further improves accuracy by fusing features and applying dual-branch refinement and calibration in both spatial and frequency domains.
\end{itemize}
\begin{itemize}
	\item Extensive experiments show that DRRNet outperforms other state-of-the-art methods on various metrics and datasets.
\end{itemize}

\section{Related Works}
\label{sec:intro}

\textbf{Camouflaged Object Detection}. COD\cite{ref1,ref2,ref3,ref4,ref5,ref6,ref7} has emerged as a particularly challenging task in computer vision due to the high similarity between the target and its background. Early approaches relied on handcrafted features such as color, texture, and edge cues to differentiate objects from complex scenes. However, these traditional methods often fell short when confronted with low contrast and ambiguous boundaries.

Thanks to the availability of large-scale datasets (e.g., CAMO\cite{ref23}, COD10K\cite{ref11}, and NC4K\cite{ref24}), researchers began to leverage the powerful representation ability of Convolutional Neural Networks to tackle COD. For instance, Fan et al. introduced SINet\cite{ref11}, which employs a two-stage framework consisting of a search module to generate candidate regions and an identification module for precise segmentation. Later, BSANet\cite{ref25} designed a residual multi-scale feature extractor to obtain multi-scale features for better understanding image content in COD tasks.

In subsequent studies, emphasis was placed on improving feature fusion and context modeling. Sun et al. developed C2FNet\cite{ref12}, which uses a multi-scale channel attention module to fuse cross-level features and enhance global context awareness. Complementarily, Zhu et al. designed BGNet\cite{ref3}, which leverages object-related edge semantics to refine boundary predictions. Additionally, data augmentation techniques were explored in MirrorNet\cite{ref26}, where horizontally flipped images provided extra cues to improve detection performance. Inspired by human visual behavior, Pang et al. adopted multi-scale zooming strategies\cite{ref27} to extract more informative cues from camouflaged targets.

More recently, Transformer-based approaches have been introduced into COD to capture long-range dependencies. Pei et al. proposed OSFormer\cite{ref28}, the first one-stage Transformer framework for COD, which integrates a reverse edge attention module to emphasize critical boundary features. Moreover, uncertainty-aware methods have been investigated to address the inherent ambiguity of camouflage scenes; for example, Kajiura et al.\cite{ref29} decoupled uncertainty reasoning and boundary estimation to achieve more robust segmentation results.

Collectively, these studies demonstrate the evolution of COD from traditional handcrafted methods to sophisticated deep architectures that integrate both CNN and Transformer components. Despite these advances, challenges such as blurred boundaries and feature misalignment persist, motivating ongoing research toward more balanced encoding-decoding strategies and dynamic feature fusion mechanisms\cite{ref30,ref31,ref32}.

\section{METHOD}
\label{sec:method}
% 在导言区引入

\subsection{Overview}
\begin{figure*}[t]
	\centering
	\includegraphics[width=1\textwidth]{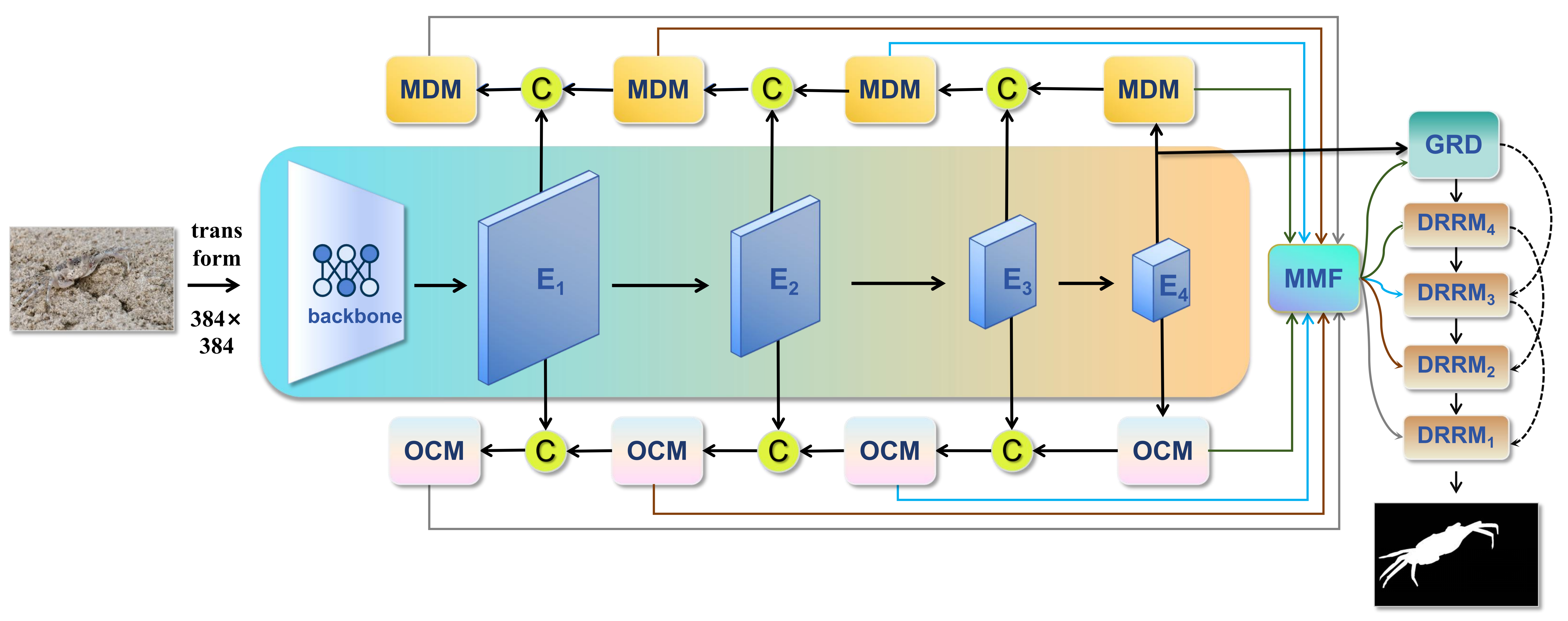}
	\caption{The framework of the proposed DRRNet.OCM extracts panoramic contextual semantics, MDM extracts local detail features, and MMF is used to fuse global and local features. GRD is responsible for generating a coarse prediction map, which serves as the prior for precise predictions. DRRM utilizes spatial edge prior information and frequency-domain noise suppression for dual reverse refinement of the results.}
	\label{fig:network_structure}
\end{figure*}
\begin{figure*}[t]
	\centering
	\includegraphics[width=1\textwidth]{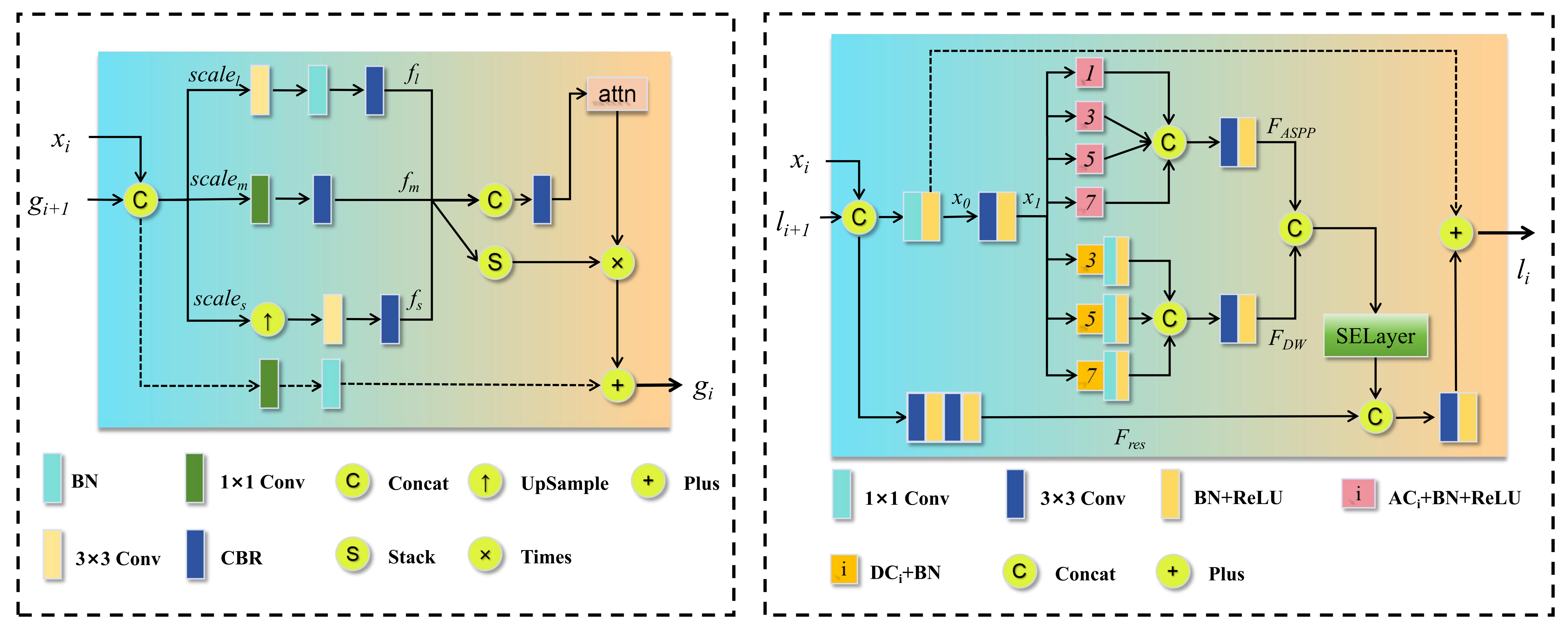}
	\caption{Details of the proposed OmniContext Module (left) and MicroDetail Module (right).}
	\label{fig:ocm_mdm_structure}
\end{figure*}

Compared with traditional methods, the proposed DRRNet adopts a four-stage framework of panoramic perception, detail mining, feature fusion, and dual reverse refinement, achieving an effective integration of global semantic information and local details which significantly improving the accuracy of camouflaged object segmentation. The overall architecture is shown in Fig.\ref{fig:network_structure}.

For an input image $I \in \mathbb{R}^{H \times W \times 3}$, the image is first fed into a backbone encoder(Pyramid Vision Transformer\cite{ref32}) to extract multi-scale feature maps:$\{x_i\}_{i=1}^N=E(I)$, where $x_i \in \mathbb{R}^{H_i \times W_i \times C_i}$. Here, $x_4$ represents the deepest features containing global semantic information, while $x_1$ corresponds to shallow features that preserve rich local details. Based on the encoder's output, the network is designed with two parallel branches: a global branch (Macro) and a local branch (Micro). The global branch processes features at each scale using the OmniContext Module (OCM) and enables information transfer between different scale features through upsampling: $g_i = \text{OCM}(\Phi(x_i, \mho(g_{i+1}))$. On the other hand, the local branch processes features at corresponding scales using the Micro Detail Module (MDM). Cross-scale information transmission is also achieved via concatenation and upsampling: $l_i = \text{MDM}(\Phi(x_i, \mho(l_{i+1})))$. Here, $\mho(\cdot)$ denotes the pixel rearrangement and upsampling operation applied to $g_{i+1}$ and $\Phi(\cdot)$ denotes the concatenation of features along the channel dimension.

To enable effective complementarity between global context and local details\cite{ref33}, we design a cross-branch fusion module, termed Macro-Micro Fusion (MMF). In this module, the outputs from the global branch $g_i$ and the local branch $l_i$ are fused. The fusion process is formulated as $f_{i} = \text{MMF}(g_i, l_i),\quad i \in \left \{ 1, 2, 3, 4 \right \}$. Finally, the fused feature $f_4$ is concatenated with the original deepest encoder feature $x_4$ and fed into a Global Rough Decoder (GRD) to generate the initial coarse prediction map $o_4 = \text{GRD}(\Phi(x_4, f_4))$. This initial prediction roughly localizes the object and outlines its shape, but fine details are still underrepresented. To further refine prediction accuracy\cite{ref34}, we introduce a series of Dual-domain Reverse Refinement Modules (DRRM), which perform progressive refinement from deep to shallow layers.

\subsection{Feature Extraction Branch}
In the encoding stage, we use a PVTv2-based backbone network\cite{ref32} $E$ to extract multi-scale features from the input image $I$. After passing through the encoder, we obtain four feature maps at different scales: $x_4, x_3, x_2, x_1$. These multi-scale features are then separately fed into the Macro Branch (global branch) and the Micro Branch (local branch).

\textbf{Macro Branch (OmniContext Module)}: To extract global semantic information, the network employs the Omni-Context Module (OCM) to process multi-scale features in a top-down manner, from deep to shallow layers. To enable cross-scale information flow, the current layer’s features are concatenated with the upsampled output from the previous (deeper) layer before being passed into the OCM. Motivated by\cite{ref34}, the OCM consists of three parallel sub-branches, each designed to capture contextual information at different receptive field scales:L-Branch applies a 3×3 convolution with stride 2 for downsampling, capturing coarse global context. M-Branch uses a 1×1 convolution to maintain original resolution and preserve mid-level features. S-Branch upsamples via bilinear interpolation, followed by a 3×3 convolution for fine-detail extraction, as shown in Fig.\ref{fig:ocm_mdm_structure}.

All sub-branch outputs are rescaled via bilinear interpolation to match the spatial resolution of the M-Branch. They are then enhanced through individual convolution operations and fused to form the final global feature representation. Specifically, the three features are concatenated along the channel dimension as follows:
\begin{equation}
	f_c = \Phi(\textit{f}_l, \textit{f}_m, \textit{f}_s) \in \mathbb{R}^{B \times 3C \times H \times W}.
\end{equation}
Next, an attention mechanism is employed to dynamically compute the weight for each scale 
$\omega \in \mathbb{R}^{B \times 3 \times 1 \times H \times W}$. The global feature is computed by performing a weighted sum over the three branches:
\begin{equation}
	g_i = \sum_{k \in \{l, m, s\}}\textit{SE}( \textit{CBR}(\textit{f}_c)) \cdot \textit{f}_k.
\end{equation}
where $SE(\cdot)$ denotes SENet\cite{ref35} self-attention module and $\textit{CBR}(\cdot)$ denotes the combination of Convolution, Batch Normalization, and ReLU activation. Finally, the residual path\cite{ref37} of the input feature $x$ is added to the fused feature to produce the final output of the OCM module: $g_i = g_i + \textit{Residual}(x)$. This multi-scale contextual modeling combined with attention fusion mechanism effectively enhances the model's global perception capability for camouflaged objects.

\textbf{Micro Branch (Micro Detail Module)}: Camouflaged object detection requires precise modeling of local details such as edges and textures, which are difficult to capture using single-layer convolutions due to their limited capacity to handle scale variations. Inspired by the work of Sun et al.\cite{ref33}, we design a dual-path structure for fine-grained feature extraction, as illustrated in Fig.\ref{fig:ocm_mdm_structure}. The first path adopts an Atrous Spatial Pyramid Pooling (ASPP)\cite{ref38} module with multiple dilation rates to aggregate multi-scale contextual information. The second path utilizes depthwise separable convolutions to extract high-resolution local features with enhanced efficiency.

For the ASPP branch, the input feature $x$ is first passed through a 1×1 convolution to adjust the channel dimensions, followed by a 3×3 convolution to extract initial local features, denoted as $x_1$. To capture local contextual information at multiple scales, the MDM module employs several parallel 3×3 convolutional branches with different dilation rates $r$. These branches generate a set of dilated features, which are then concatenated along the channel dimension and passed through a convolution layer for dimensionality reduction, yielding the final output of the ASPP branch $F_{aspp}$:
\begin{equation}
	F_{aspp} = \textit{conv}\Bigl(\Phi\bigl(\{f_{aspp}^{(r)}\}_{r\in\{1,3,5,7\}}\bigr)\Bigr).
\end{equation}

where $\textit{conv}(\cdot)$ denotes as a sequential of convolution operation. For the depthwise separable convolution branch, 3×3, 5×5, and 7×7 depthwise separable convolutions are applied to the input feature map to obtain three parallel feature representations. These features are then concatenated along the channel dimension and passed through a convolution layer for dimensionality reduction, resulting in the output of the depthwise separable branch $F_{dw}$:

\begin{equation}
	F_{dw} = \text{con}\Bigl(\Phi\bigl(f_{dw}^{(3)}, f_{dw}^{(5)}, f_{dw}^{(7)}\bigr)\Bigr).
\end{equation}

To fully leverage the information from both the ASPP branch and the depthwise separable convolution branch, the module concatenates their outputs along the channel dimension. The concatenated features are then modulated using an SENet\cite{ref35} attention module, resulting in the local detail features $F_{local}$. At the same time, to complement the original details in the input, a residual branch\cite{ref37} performs additional convolution on the input, denoted as $F_{res}$. Finally, 
$F_{local}$ and $F_{res}$ are concatenated along the channel dimension, followed by fusion convolution, and then the initially adjusted features $x_{0}$ are added to form the final local feature output: 
\begin{equation}
	l_{i} = \textit{CBR}\Bigl(\Phi(\textit{SE}\Bigl(\text{cat}(F_{aspp}, F_{dw}), F_{res})\Bigr) + x_0.
\end{equation}
The local feature $l_i$ can effectively enhance the model's ability to perceive fine details of the target.

\subsection{Feature Fusion Branch}
\begin{figure}[htbp]
	\centering
	\includegraphics[width=1\linewidth]{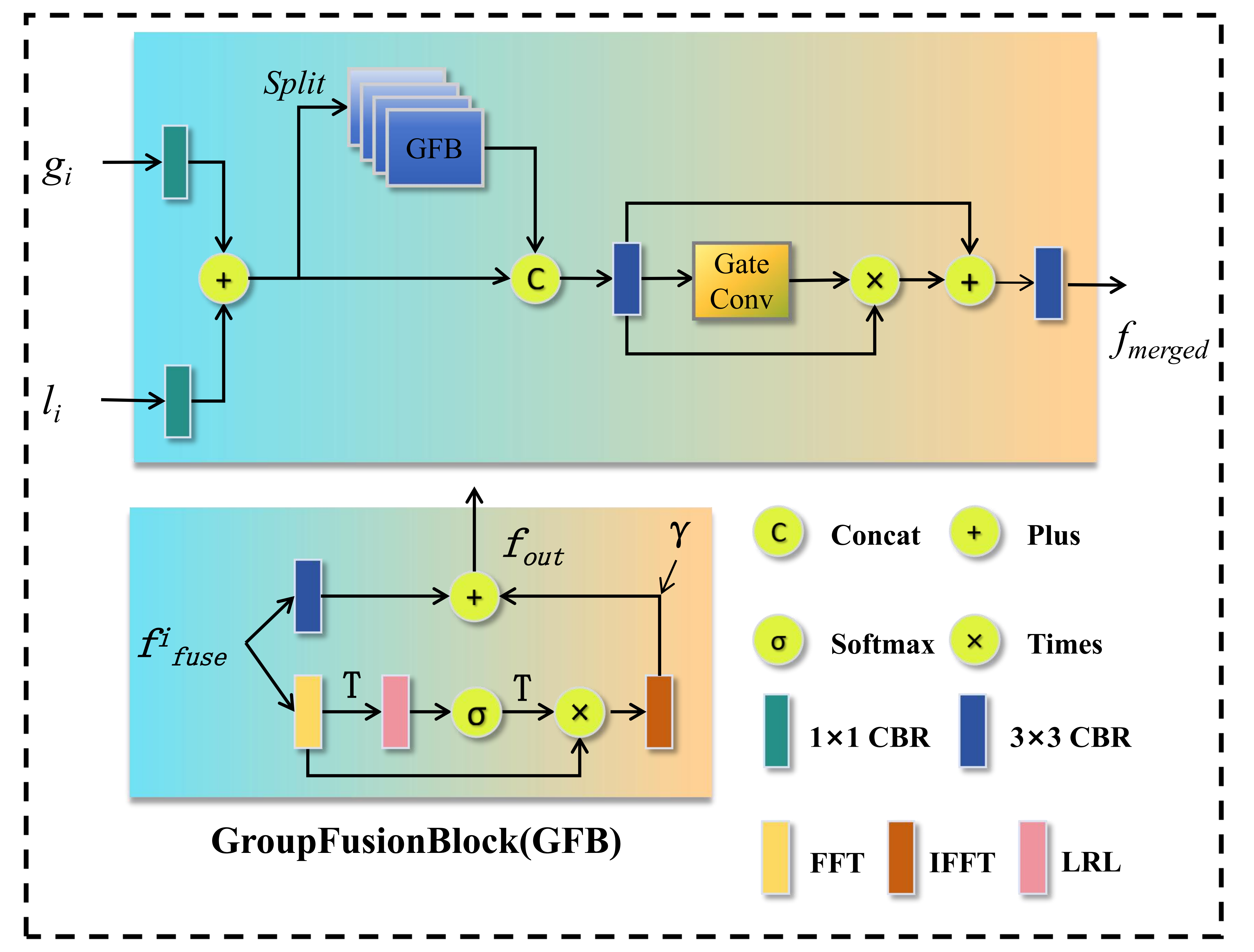}
	\caption{The details of Macro-Micro Fusion(MMF) module.}
	\label{fig:mmf_structure}
\end{figure}

Given a global semantic feature $g_i$ and a local detail feature $l_i$, we propose a Group-wise Mixed Fusion Module to dynamically integrate the two streams\cite{ref39}, enabling mutual complementarity and information enhancement. Specifically, we first compute the preliminary fused feature by channels's concat of $g_i$ and $l_i$, and then divide the result into four channel-wise groups, each with C/2 channels, as shown in Fig.\ref{fig:mmf_structure}. For each group $f_i$, we adopt a GroupFusionBlock to perform joint modeling of spatial and frequency representations. The outputs of all groups are concatenated along the channel dimension and fused with the initial fused feature via gated convolution\cite{ref40} to filter redundant information and enhance feature discriminability. A residual connection\cite{ref37} is subsequently employed to produce the final fused representation $F_i$.

In the GroupFusionBlock, we utilize both a spatial branch and a frequency branch\cite{ref41} for dual-path modeling. The spatial branch captures local structure via standard 3×3 convolution and outputs $x_{spatial}$. For the frequency branch, we first apply 2D Fast Fourier Transform (FFT)\cite{ref42} to the input $x$ to obtain complex-domain representations $X$. The amplitude of $X$ is fed into a lightweight MLP to generate channel-wise modulation weights, which are then used to modulate the original $X$. After modulation, the inverse FFT (IFFT) is applied to transform the result back to the spatial domain, yielding the frequency-enhanced feature $x_{freq}$. The final fused output for each group is computed by a learnable weighted summation of the two branches:

\begin{equation}
	\begin{aligned}
		&x_{\text{spatial}} = \textit{CBR}(\Phi(g_{i}, l_{i})), \quad i \in \{1,2,3,4\} \\
		&x_{\text{freq}} = \mathcal{F}^{-1}(\mathcal{F}(x) \cdot \Psi (|\mathcal{F}(x)|) \\
		&f^i_{fused} = x_{\text{spatial}} + \gamma_{\text{GFB}} \cdot x_{\text{freq}}.\\
		&F_{\text{fused}} = \Phi \left(_{k=1}^{4}\Bigl(f^{(k)}_{\text{fused}}\Bigr)\right).\\
		&F_{\textit{i}} = \gamma \cdot \theta\Bigl(F_{\text{fused}}\Bigr) \times F_{\text{fused}} + F_{\text{fused}}.\\
	\end{aligned}
\end{equation}
where $\textit{CBR}(\cdot)$ denotes a sequential combination of Convolution, Batch Normalization, and ReLU activation.$\mathcal{F}(\cdot)$ and $\mathcal{F}^{-1}(\cdot)$ represent the forward and inverse Fourier transforms, respectively.$\Psi(\cdot)$ refers to a Sigmoid-activated MLP used for generating adaptive modulation weights.$\gamma_{\text{GFB}}$ is a learnable scalar that controls the contribution of the frequency-aware branch. $\theta(\cdot)$denotes a gated convolutional operation. The fused features integrate both global semantic understanding and local detail awareness, thereby enhancing the detection capability for camouflaged objects.

%-------------------------------------------------------------------------
\subsection{Feature Decoding Branch}
\subsubsection{Global Rough Decoder Module}
Although the encoder extracts multi-scale representations via global and local branches, the resulting fused feature $F_i$ still resides in a high-dimensional feature space, with a resolution significantly lower than that of the input image, making it unsuitable for direct pixel-level prediction. To address this, we introduce a Global Rough Decoder (GRD) that transforms the deep fused features into a coarse prediction map, providing global priors for subsequent fine-grained segmentation.

As shown in Fig.~\ref{fig:ard_structure}, GRD is composed of three main components: a shallow convolutional stack for feature enhancement, a novel Multi-Scale Attention (MSA) module for context modeling, and a lightweight segmentation head for coarse prediction. Given the deepest fused feature, we first obtain two base features $p_{s0}$ and $p_{s1}$ through sequential 1×1 and 3×3 convolutional layers. This base feature is then fed into the MSA module, which is our custom-designed attention block that simultaneously captures local details and large-range dependencies.We employ multi-branch dilated convolutions to extract features at various receptive field scales. These features are subsequently aggregated and adaptively enhanced through a SENet attention mechanism, resulting in a context-aware representation.The formula is as follows:
\begin{figure}[htbp]
	\centering
	\includegraphics[width=1\linewidth]{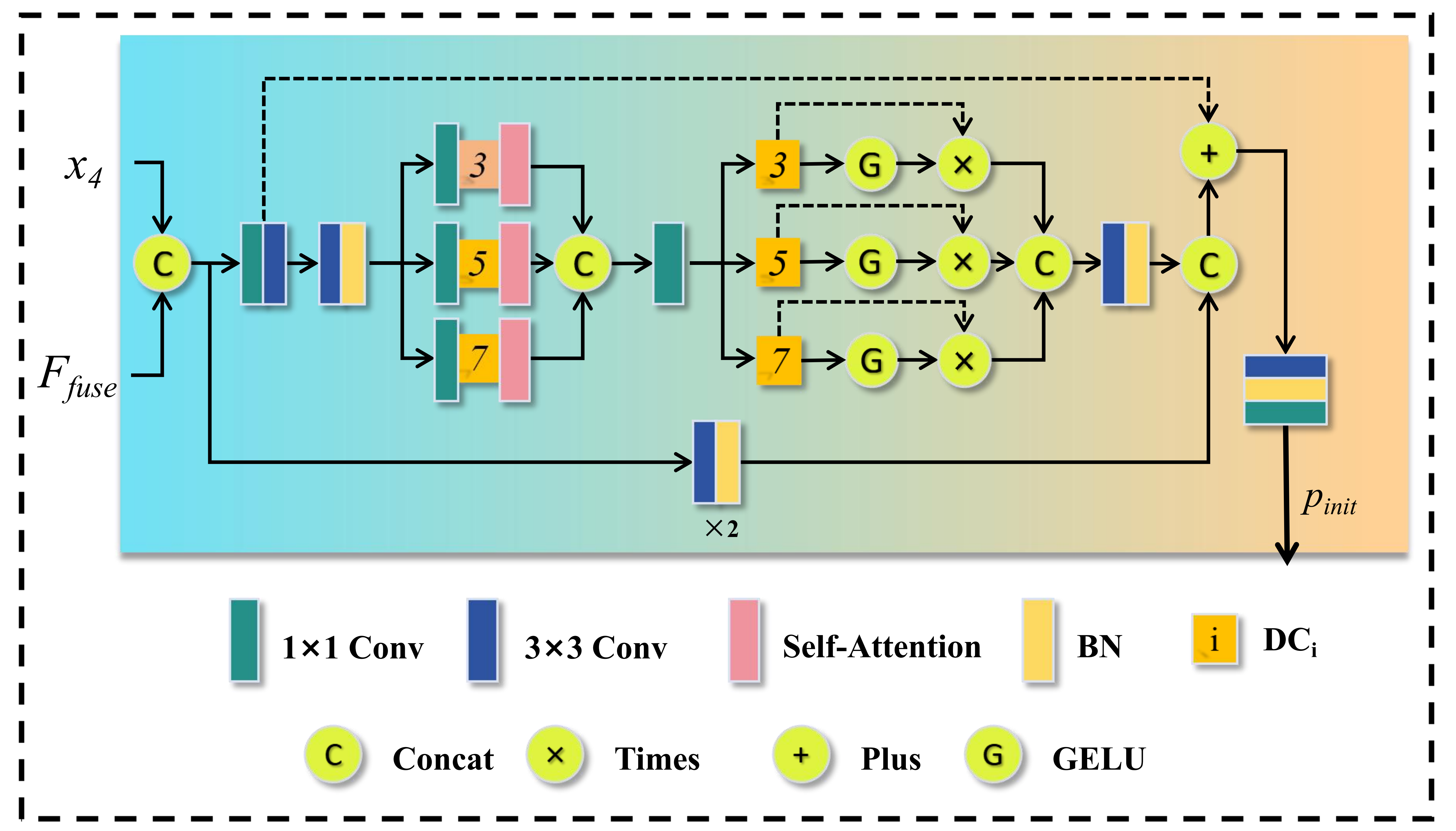}
	\caption{The details of Global Rough Decoder(GRD) module.}
	\label{fig:ard_structure}
\end{figure}

\begin{equation}
	\begin{aligned}
		&p_{s2} = \Phi \left ( \textit{SE}\Bigl(\textit{DW}\bigl(\{p^{i}_{s1}\}_{i\in\{3,5,7\}}\bigr)\Bigr) \right ) \\
		&p^i_{s3} = \textit{DW}\bigl(\{p_{s2}^{(i)}\}_{i\in\{3,5,7\}}\bigr). \\
		&p_{s3} = \Phi \left (\{ \Omega \left (  p^i_{s3}\right ) \odot  p^i_{s3}\}_{i\in\{3,5,7\}}\right ).\\
		&O_4 = conv(\Phi \left( p_{s0},p_{s3} \right) + p_{s1})\\
	\end{aligned}
\end{equation}

where $\textit{DW}(\cdot)$ denotes the depthwise separable convolution\cite{ref43}. $\textit{SE}(\cdot)$ denotes the SE attention module\cite{ref35}. $\Phi(\cdot)$ denotes the concatenation of features along the channel dimension. $\Omega(\cdot)$ denotes the Gaussian Error Linear Unit (GELU) activation function\cite{ref44}. $\odot$ denotes element-wise multiplication.

\subsubsection{Dual Reverse Refinement Module}

\begin{figure}[htbp]
	\centering
	\includegraphics[width=1\linewidth]{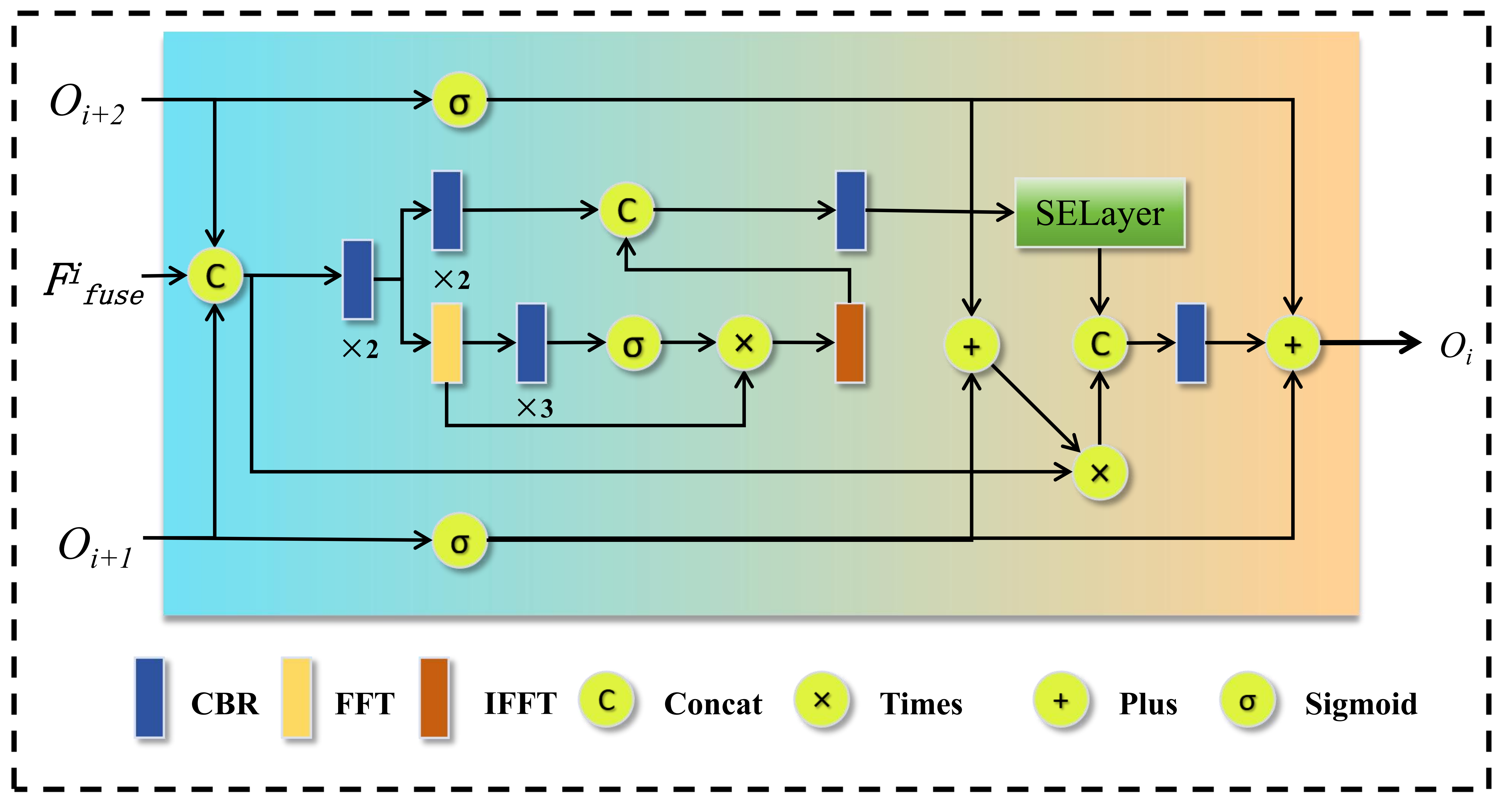}
	\caption{The details of Dual Reverse Refinement Module(DRRM).}
	\label{fig:drrm_structure}
\end{figure}

In our proposed DRRM, we refine the fused features by integrating multi-level contextual information while simultaneously considering spatial and frequency domain characteristics. Specifically, feature maps from different layers, namely $O_{i+1}$, $O_{i+2}$ and the fused feature $F_i$, are mapped into the DRRM to generate the refined feature map $O_i$, as illustrated in Fig.\ref{fig:drrm_structure}. Technically, $O_{i+1}$ and $O_{i+2}$ are first upsampled and dimensionally expanded to match the channel and spatial dimensions of $F_i$ before concatenation. This process yields a preliminary fused feature $F_c$, which is computed as $F_{c} = \Phi (F^i_{fuse}, \mho (O_{i+1}), \mho (O_{i+2}))$. where $\mho(\cdot)$ denotes the upsampling and dimension expansion operation, and $\Phi(\cdot)$ represents the channel-wise concatenation operation.

Subsequently, $F_c$ is fed into two parallel refinement branches. In the spatial branch, depthwise separable convolutions\cite{ref43} are employed to extract local details, resulting in the refined spatial feature $F_s = CBR\left(F_c\right)$, where CBR represents Conv+BN+ReLU. In the frequency branch, we first compute the Fourier transform\cite{ref42} of $F_c$ and extract its real part. After processing this through a convolutional layer to generate modulation weights, the modulated frequency features are transformed back to the spatial domain using the inverse Fourier transform, yielding $F_f$, which is computed as $F_{f} = \mathcal{F}^{-1}\big(Conv(\Re(\tilde{F})(\mathcal{F}(c)))\odot \mathcal{F}(c)\big)$. where $\mathcal{F}(c)$ and $\mathcal{F}^{-1}$ denotes as the Fourier transform and inverse Fourier transform. $\Re(\tilde{F})$ denotes the extraction of the real part of the Fourier transform, and $\odot$ denotes element-wise multiplication.The spatial feature $F_c$ and the frequency feature $F_s$ are then concatenated and passed through an additional convolutional layer, after which a SENet\cite{ref35} module is applied to generate the attention feature $F_{attn}=\textit{SE}(conv(\Phi(F_c, F_s))$. This SE module adaptively re-calibrates the channel-wise responses by emphasizing critical features and suppressing redundant information.

For the prior residual, outputs $O_{i+1}$ and $O_{i+2}$ are each processed by a sigmoid activation followed by an inversion, producing residual maps $R_1$ and $R_2$.These residual maps are summed and then element-wise multiplied with $F_c$ to produce the weighted residual feature $F_{w}$. Finally, the attention feature $F_{attn}$ and the weighted residual $F_{w}$ are fused via $\Phi(\cdot)$ and processed by a convolutional layer. A residual connection\cite{ref37} from $O_{i+1}$ and $O_{i+2}$ is then added to yield the final output $O_i$.This process can be expressed as:
\begin{equation}
	\begin{aligned}
		R_1 &= 1 - \sigma(O_{i+1}), \quad R_2 = 1 - \sigma(O_{i+2}), \\
		F_{\text{w}} &= (R_1 + R_2) \odot F_{\text{c}},\\
		O_i& = conv(\Phi(F_{attn}, F_{w})) + O_{i+1} + O_{i+2}\\
	\end{aligned}
\end{equation}
where $\sigma(\cdot)$ denotes the sigmoid function, $\odot$ represents element-wise multiplication, and $\Phi(\cdot)$ denotes channel-wise concatenation.

\subsection{Loss Function}
In supervised training, we draw on the methods of references\cite{ref25,ref33,ref45,ref46}, and with the help of ground truths, combine the weighted binary cross - entropy loss ($\zeta_{BCE}^w$) and the weighted intersection - over - union loss ($\zeta_{IoU}^w$). The expression of the loss function $\zeta_{all}$ is:
\begin{equation}
	\zeta_{all} = \sum_{i = 0}^4 \left( \zeta_{BCE}^w (\sigma(O_i), G) + \zeta_{IoU}^w (\sigma(O_i), G) \right)
\end{equation}
where $\sigma(\cdot)$ represents the sigmoid function and G represents the ground truth.

Intersection-over-union. union (IoU) loss\cite{ref47} is used to calculate the complete structural similarity. The definition of $\zeta_{IoU}^w$ is:
\begin{equation}
	\zeta_{IoU}^w = 1 - \frac{\sum_{(i,j)}^{HW} (1 + w_{ij}) G_{ij} P_{ij}}{\sum_{(i,j)}^{HW} (1 + w_{ij}) \left( G_{ij} + P_{ij} - G_{ij} P_{ij} \right)}
\end{equation}
where \(w_{ij}\) is the weight value of the pixel \((i, j)\), and \(P_{ij}\) and \(G_{ij}\) represent the predicted value and the ground - truth value of the pixel \((i, j)\) as a camouflaged object, respectively.

\section{Experiments}
\begin{table*}[!ht]
	\centering
	\small
	\renewcommand{\arraystretch}{1.2}
	\setlength{\tabcolsep}{4.5pt}
	\caption{Performance comparison with baseline models on COD datasets. $\uparrow$ and $\downarrow$ represent that the higher and the lower the better respectively. Red is the best result.}
	\label{tab:sota_comparison}
	\begin{tabular}{l|cccc|cccc|cccc}
		\hline
		\multirow{2}{*}{Methods}  & \multicolumn{4}{c|}{CAMO} & \multicolumn{4}{c|}{COD10K} & \multicolumn{4}{c}{NC4K} \\
		& $S_\alpha\uparrow$ & $E_\phi\uparrow$ & $F^{\omega}_\beta\uparrow$ & $M\downarrow$ 
		& $S_\alpha\uparrow$ & $E_\phi\uparrow$ & $F^{\omega}_\beta\uparrow$ & $M\downarrow$ 
		& $S_\alpha\uparrow$ & $E_\phi\uparrow$ & $F^{\omega}_\beta\uparrow$ & $M\downarrow$ \\
		\hline
		
		FEDER\textsubscript{23\_CVPR}\cite{ref53}  & 0.836 & 0.897 & 0.807 & 0.066 
		& 0.844 & 0.911 & 0.748 & 0.029 
		& 0.862 & 0.913 & 0.824 & 0.042 \\
		
		EAMNet\textsubscript{23\_ICME}\cite{ref54} & 0.831 & 0.890 & 0.763 & 0.064 
		& 0.839 & 0.907 & 0.733 & 0.029 
		& 0.862 & 0.916 & 0.801 & 0.040 \\
		
		DGNet\textsubscript{23\_MIR}\cite{ref55}  & 0.855 & 0.912 & 0.795 & 0.042 
		& 0.822 & 0.877 & 0.693 & 0.033 
		& 0.857 & 0.907 & 0.784 & 0.042 \\
		
		PENet\textsubscript{23\_IJCAI}\cite{ref56}  & 0.828 & 0.890 & 0.771 & 0.063 
		& 0.831 & 0.908 & 0.723 & 0.031 
		& 0.855 & 0.912 & 0.795 & 0.042 \\
		
		OPNet\textsubscript{23\_IJCV}\cite{ref57}  & 0.858 & 0.915 & 0.817 & 0.050 
		& 0.857 & 0.919 & 0.767 & 0.026 
		& 0.883 & 0.932 & 0.838 & 0.034 \\
		
		FSPNet\textsubscript{23\_CVPR}\cite{ref45} & 0.856 & 0.899 & 0.799 & 0.050 
		& 0.851 & 0.895 & 0.735 & 0.026 
		& 0.879 & 0.915 & 0.816 & 0.035 \\
		
		MSCAF-Net\textsubscript{23\_TCSVT}\cite{ref59} & 0.873 & 0.929 & 0.828 & 0.046 
		& 0.865 & 0.927 & 0.775 & 0.024 
		& 0.887 & 0.935 & 0.838 & 0.032 \\
		
		SDRNet\textsubscript{24\_KBS}\cite{ref20}  & 0.872 & 0.924 & 0.826 & 0.049 
		& 0.871 & 0.924 & 0.785 & 0.023 
		& 0.889 & 0.934 & 0.842 & 0.032 \\
		
		UEDG\textsubscript{24\_TMM}\cite{ref58}  & 0.868 & 0.922 & 0.819 & 0.048 
		& 0.858 & 0.924 & 0.766 & 0.025 
		& 0.881 & 0.928 & 0.829 & 0.035 \\
		
		CMNet\textsubscript{24\_TCSVT}\cite{ref63} & 0.835 & 0.902 & 0.828 & 0.063 
		& 0.834 & 0.912 & 0.778 & 0.030 
		& 0.859 & 0.922 & 0.843 & 0.041 \\
		
		DRFNet\textsubscript{24\_TCSVT}\cite{ref62} & 0.868 & 0.925 & 0.832 & 0.047 
		& 0.869 & 0.936 & 0.792 & 0.023 
		& 0.887 & 0.939 & 0.846 & 0.031 \\
		
		MVGNet\textsubscript{24\_TCSVT}\cite{ref61} & 0.879 & 0.930 & 0.839 & 0.045 
		& 0.877 & 0.936 & 0.799 & 0.022 
		& 0.894 & 0.938 & 0.850 & 0.030 \\
		
		AGLNet\textsubscript{24\_arXiv}\cite{ref66}  & 0.873 & 0.919 & 0.826 & 0.049 
		& 0.871 & 0.927 & 0.792 & 0.023 
		& 0.887 & 0.933 & 0.838 & 0.033 \\
		
		BDCL-Net\textsubscript{25\_KBS}\cite{ref60} & 0.881 & 0.929 & 0.845 & \textcolor{red}{\textbf{0.040}} 
		& 0.869 & 0.935 & 0.790 & 0.022 
		& 0.888 & 0.932 & 0.844 & 0.032 \\
		
		FLRNet-P\textsubscript{25\_NeuCom} & 0.877 & 0.927 & 0.833 & 0.045 
		& 0.868 & 0.929 & 0.781 & 0.024 
		& 0.890 & 0.935 & 0.842 & 0.032 \\
		
		EFNet\textsubscript{25\_NeuCom}\cite{ref21}  & 0.881 & 0.930 & 0.847 & 0.041 
		& 0.880 & 0.936 & 0.805 & 0.021 
		& 0.896 & 0.941 & 0.856 & 0.029 \\
		
		PRBENet\textsubscript{25\_TMM}\cite{ref65}  & 0.876 & 0.928 & 0.837 & 0.045 
		& 0.867 & 0.932 & 0.793 & 0.021 
		& 0.887 & 0.931 & 0.845 & 0.031 \\
		
		\textbf{DRRNet(Ours)} & \textcolor{red}{\textbf{0.881}} & \textcolor{red}{\textbf{0.934}} & \textcolor{red}{\textbf{0.848}} & 0.041 
		& \textcolor{red}{\textbf{0.881}} & \textcolor{red}{\textbf{0.937}} & \textcolor{red}{\textbf{0.817}} & \textcolor{red}{\textbf{0.019}} 
		& \textcolor{red}{\textbf{0.896}} & \textcolor{red}{\textbf{0.941}} & \textcolor{red}{\textbf{0.859}} & \textcolor{red}{\textbf{0.028}} \\
		
		\hline
	\end{tabular}
\end{table*}

\begin{table*}[ht]
	\centering
	\small
	\renewcommand{\arraystretch}{1.2}
	\setlength{\tabcolsep}{3pt}
	\caption{Ablation study on CAMO and NC4K datasets with different module combinations.}
	%\rowcolors{2}{blue!5}{white} % 开启交错颜色

	\begin{tabular}{c|ccccc|cccc|cccc}
		\hline
		
		\multirow{2}{*}{\textbf{Baseline}} 
		& \multicolumn{5}{c|}{\textbf{Module}} 
		& \multicolumn{4}{c|}{\textbf{CAMO (250 images)}} 
		& \multicolumn{4}{c}{\textbf{NC4K (4121 images)}} \\
		& \textbf{OCM} & \textbf{MDM} & \textbf{MMF} & \textbf{GRD} & \textbf{DRRM} 
		
		& $M\downarrow$ & $F^{\omega}_\beta\uparrow$ & $S_\alpha\uparrow$ & $E_\phi\uparrow$  
		& $M\downarrow$ & $F^{\omega}_\beta\uparrow$ & $S_\alpha\uparrow$ & $E_\phi\uparrow$  \\
		\hline

		%Base
		\checkmark & & & & & & 0.0701 & 0.8020 & 0.8028 & 0.8648 & 0.0451 & 0.8131  & 0.8297 & 0.8906 \\
		
		%Base+OCM
		\checkmark & \checkmark  &  & & & & 0.0534 & 0.8215 & 0.8344 & 0.8982 & 0.0370 & 0.8524  & 0.8599 & 0.9159 \\
		%Base+MDM
		\checkmark &  & \checkmark &  & & & 0.0571 & 0.8312 & 0.8429 & 0.9006 & 0.0391 & 0.8530 & 0.8647 & 0.9146 \\
		%Base+MMF
		%\checkmark &  &   & \checkmark & & & 0.078 & 0.761 & 0.714 & 0.801 & 0.868 & 0.039 & 0.681 & 0.661 & 0.801 & 0.863 \\
		
		%Base+GRD
		%\checkmark &  &   &  & \checkmark & & 0.078 & 0.761 & 0.714 & 0.801 & 0.868 & 0.039 & 0.681 & 0.661 & 0.801 & 0.863 \\	
		
		%Base+DRRM
		%\checkmark &  &   &  &  & \checkmark & 0.078 & 0.761 & 0.714 & 0.801 & 0.868 & 0.039 & 0.681 & 0.661 & 0.801 & 0.863 \\
		
		%Base+OCM+MDM
		\checkmark & \checkmark& \checkmark & & & & 0.0464 & 0.8440& 0.8533 & 0.9179 & 0.0357 & 0.8603 & 0.8697 & 0.9192 \\

		%Base+OCM+MDM+MMF
		\checkmark & \checkmark & \checkmark & \checkmark & & & 0.0457 & 0.8645 &  0.8615 & 0.9272 & 0.0341 & 0.8815 & 0.8790 & 0.9306 \\
		
		\checkmark &  \checkmark & \checkmark & \checkmark & \checkmark & & 0.0445 & 0.8765 & 0.8717 & 0.9320 & 0.0312 & 0.8864 & 0.8832 & 0.9385 \\

		\checkmark & \checkmark & \checkmark & \checkmark & & \checkmark &   0.0434 & 0.8785 & 0.8760 & 0.9343 & 0.0312 & 0.8913 & 0.8868 & 0.9358 \\

		\checkmark & \checkmark &\checkmark & \checkmark & \checkmark & \checkmark & \textcolor{red}{\textbf{0.0415}} & \textcolor{red}{\textbf{0.8906}} &  \textcolor{red}{\textbf{0.8819}} & \textcolor{red}{\textbf{0.9347}} & \textcolor{red}{\textbf{0.0283}} & \textcolor{red}{\textbf{0.9018}} &  \textcolor{red}{\textbf{0.8959}} & \textcolor{red}{\textbf{0.9417}} \\
		\bottomrule
	\end{tabular}
	
	\label{tab:ablation}
\end{table*}

\subsection{Experimental Setup}
$\textbf{Datasets}$. We conduct experiments on three widely used benchmark datasets of COD task, i.e., CAMO\cite{ref23}, COD10K\cite{ref11} and NC4K\cite{ref24}. In particular, CAMO, covering eight categories, contains 1,250 camouflaged images and 1,250 noncamouflaged images. COD10K consists of 5,066 camouflaged, 1,934 non-camouflaged, and 3,000 background images, and it is currently the largest dataset which covers 10 superclasses and 78 subclasses. NC4K is a newly published dataset which has a total of 4,121 camouflaged images. Following standard practice of COD tasks, we use 3,040 images from COD10K and 1,000 images from CAMO as the training set and the remaining data as the test set.

$\textbf{Evaluation Metrics}$. We employed four commonly used evaluation metrics in COD. $S_{\alpha}$\cite{ref48} for structural similarity evaluation, $E_{\phi}$\cite{ref49} for considering global and local pixel matching, $F^{\omega}_{\beta}$\cite{ref50} for balancing precision and recall, and $MAE$ for calculating the average absolute error between pixels. Higher $S_{\alpha}$, $E_{\phi}$, $F^{\omega}_{\beta}$ and lower $MAE$ indicates better segmentation performance.

$\textbf{Implementation Details}$. The proposed DRRNet is implemented with Pytorch. To achieve optimal detection performance, PVTv2 pre-trained on ImageNet\cite{ref51} is employed as the backbone. And we use Adam\cite{ref52} with an initial learning rate of 1e-4 as the optimizer and decay rate of 0.1 for every 25 epochs. Batch size is set to 8 during training and the whole network is iterated 80 epochs. Note that all input images are resized as 384×384 during the training and testing process. Additionally, similar to SINet\cite{ref11}, BSANet\cite{ref25}, and FSPNet\cite{ref45}, we employ several data augmentation strategies (e.g., horizontal flipping, random cropping, and color enhancement) during the training process to prevent model overfitting.

\subsection{Comparison with SOTA methods}

\begin{figure*}[t]
	\centering
	\includegraphics[width=0.9\textwidth]{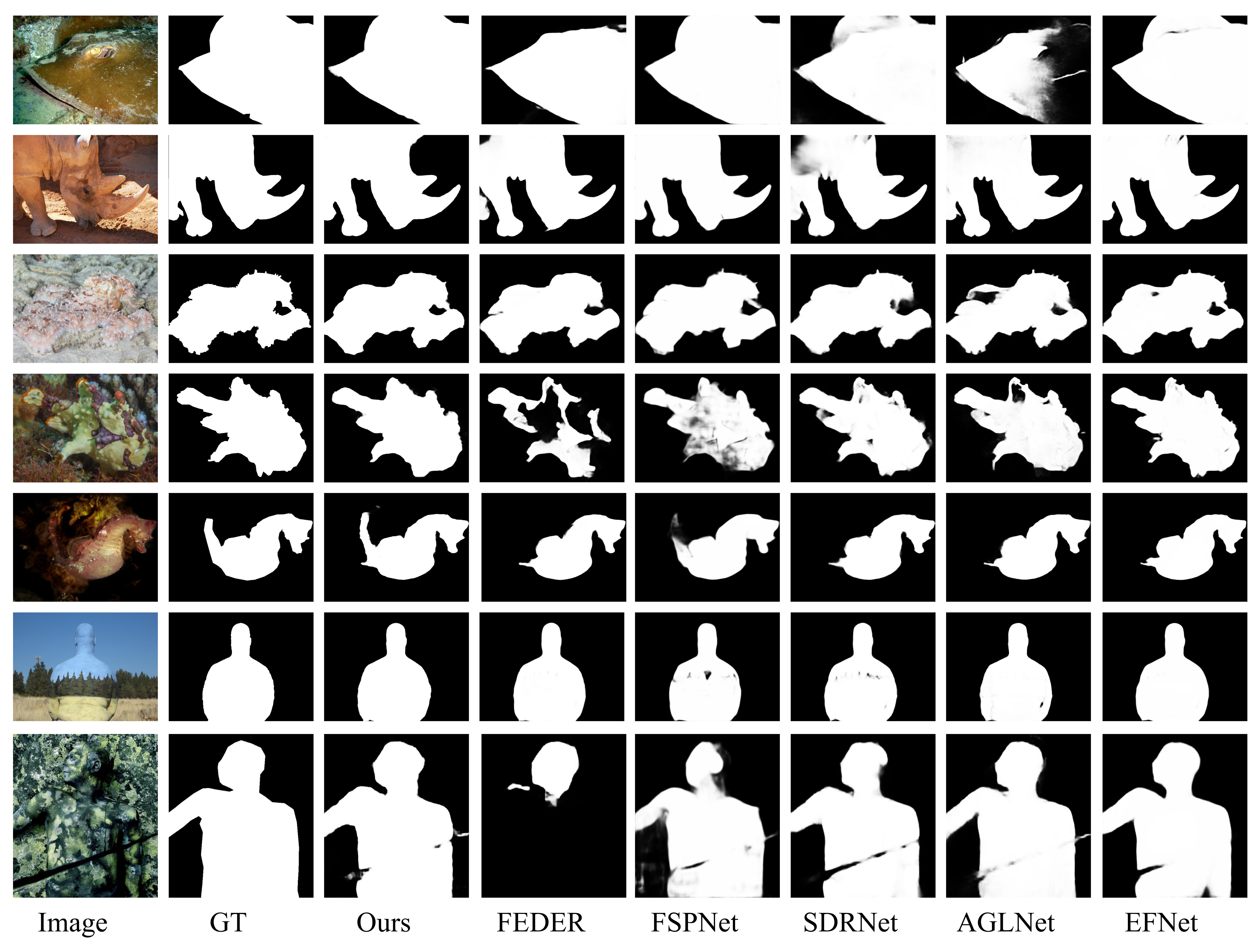}
	\caption{The visualization comparisons between our DRRNet and other stronger SOTA methods in various scenarios. Under the conditions of target size variation, complex scenes (occlusion, micro details), and artificial camouflaged detection, the proposed DRRNet is significantly better than the compared sota methods in the integrity of target recognition and suppression of false recognition.}
	\label{fig:compare_sota}
\end{figure*}

\textbf{Quantitative Evaluation}. To validate the effectiveness of the proposed DRRNet, we compare it against 17 state-of-the-art (SOTA) methods. For fairness, all prediction maps are generated using identical code, and all metrics are computed with the same implementation. Detailed comparisons are presented in Table.~\ref{tab:sota_comparison}. Notably, DRRNet significantly outperforms all SOTA methods across all datasets without relying on any post-processing or specialized techniques. These quantitative results highlight the effectiveness of our network, demonstrating the superiority of the proposed approach in addressing the challenging task of COD.

\textbf{Qualitative Comparison}. Fig.~\ref{fig:compare_sota} illustrates visual comparisons between our DRRNet and existing COD methods across diverse scenarios, including objects of varying sizes, camouflaged targets, and occluded instances. As shown in Fig.~\ref{fig:compare_sota}, several SOTA competitors (e.g., SDRNet\cite{ref20}, FSPNet\cite{ref45} and our lab's previous work EFNet\cite{ref21}) fail to fully segment camouflaged and occluded objects with irregular sizes. By contrast, due to the collaborative inter-module refinement in our design, DRRNet demonstrates heightened sensitivity to camouflaged objects of all scales and effectively segments targets across heterogeneous scenes. Notably, Fig.~\ref{fig:compare_sota} further reveals that DRRNet achieves superior recognition accuracy for camouflaged targets compared to state-of-the-art alternatives, particularly in cluttered backgrounds or ambiguous boundary scenarios. These visual results validate the proposed method’s ability to handle the core challenges of COD (scale variation, camouflage, occlusion) through explicit structural design.

\subsection{Ablation studies}

\subsubsection{Effectiveness of modules}

\begin{figure*}[t]
	\centering
	\includegraphics[width=0.9\textwidth]{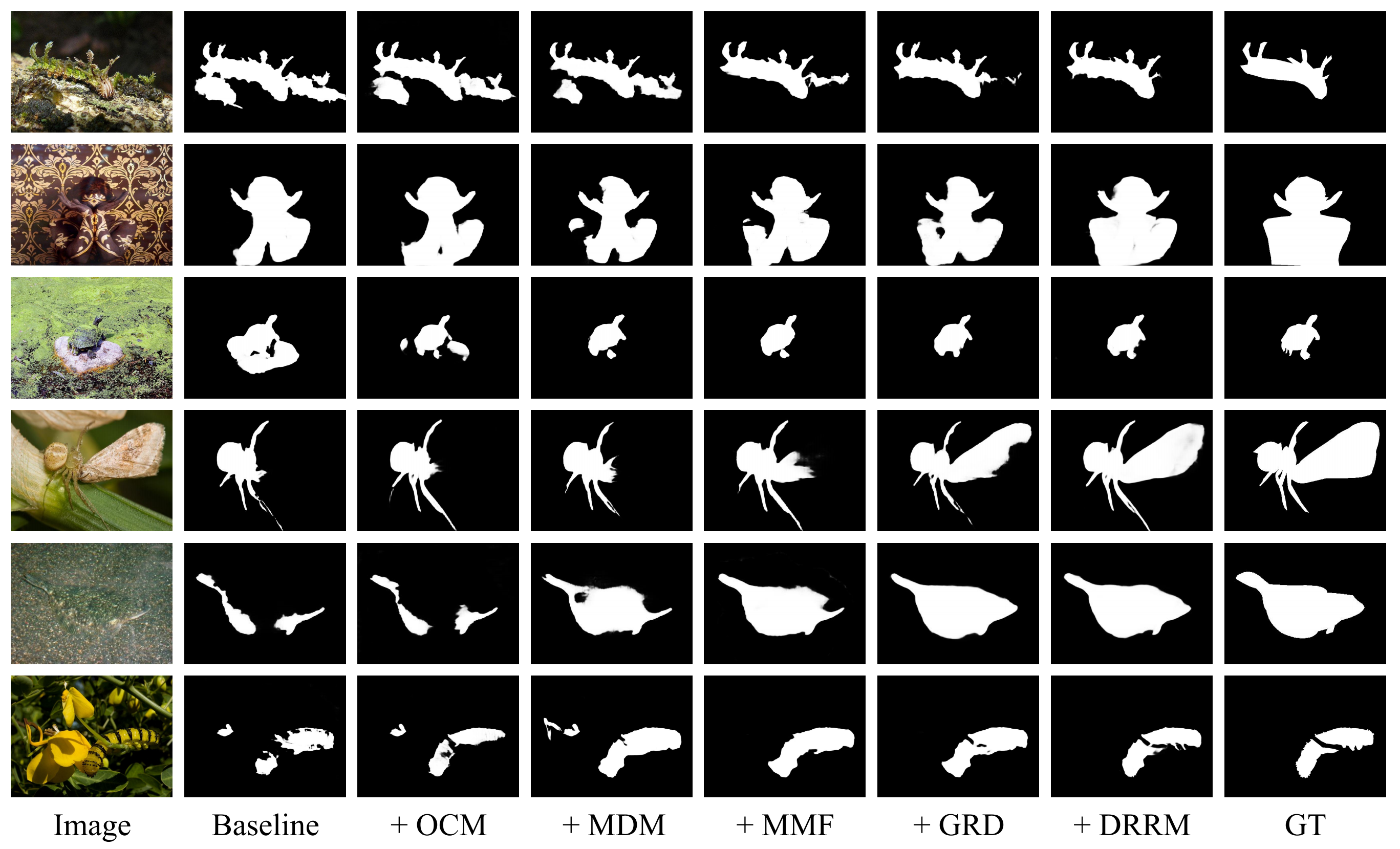}
	\caption{Visual results of the effectiveness of our modules.}
	\label{fig:ablation_module}
\end{figure*}

\begin{table*}[htbp]
	\centering
	\renewcommand{\arraystretch}{1.2}
	\setlength{\tabcolsep}{3pt}
	\footnotesize
	\caption{Ablation analysis of our OCM module.}
	\label{tab:edit_ocm}
	\begin{tabular}{ccc|cccc|cccc|cccc}
		\toprule
		\multicolumn{3}{c|}{\textbf{Structure settings}} 
		& \multicolumn{4}{c|}{\textbf{CAMO (250 images)}} 
		& \multicolumn{4}{c}{\textbf{COD10K (2026 images)}} 
		& \multicolumn{4}{c}{\textbf{NC4K (4121 images)}} \\
		
		\textbf{Baseline} & \textbf{OCM Add} & \textbf{OCM Cat} 
		& $M\downarrow$ & $F^{\omega}_\beta\uparrow$ & $S_\alpha\uparrow$ & $E_\phi\uparrow$  
		& $M\downarrow$ & $F^{\omega}_\beta\uparrow$ & $S_\alpha\uparrow$ & $E_\phi\uparrow$
		& $M\downarrow$ & $F^{\omega}_\beta\uparrow$ & $S_\alpha\uparrow$ & $E_\phi\uparrow$\\
		\midrule
		\checkmark & \checkmark & 
		& 0.0423 & 0.8473                          
		& 0.8805 & 0.9357                 
		& 0.0215                 & 0.8038
		& 0.8733 & 0.9305 
		& 0.0305 & 0.8513                      
		& 0.8903                 & 0.9384\\
		
		\checkmark &           & \checkmark 
		
		& \textcolor{red}{0.0415} & \textcolor{red}{0.8481} 
		& \textcolor{red}{0.8819} & \textcolor{red}{0.9347} 
		& \textcolor{red}{0.0196} & \textcolor{red}{0.8174}                      
		& \textcolor{red}{0.8815} & \textcolor{red}{0.9371} 
		& \textcolor{red}{0.0283} & \textcolor{red}{0.8589}                  
		& \textcolor{red}{0.8959} & \textcolor{red}{0.9417}\\
		\bottomrule
	\end{tabular}
	
\end{table*}

\begin{table*}[!htbp]
	\centering
	\renewcommand{\arraystretch}{1.2}
	\setlength{\tabcolsep}{3pt}
	\footnotesize
	\caption{Ablation analysis of our MDM module.}
	\label{tab:edit_MDM}
	\begin{tabular}{ccc|cccc|cccc|cccc}
		\toprule
		\multicolumn{3}{c|}{\textbf{Structure settings}} 
		& \multicolumn{4}{c|}{\textbf{CAMO (250 images)}} 
		& \multicolumn{4}{c}{\textbf{COD10K (2026 images)}} 
		& \multicolumn{4}{c}{\textbf{NC4K (4121 images)}} \\
		
		\textbf{Baseline} & \textbf{MDM Add} & \textbf{MDM Cat} 
		& $M\downarrow$ & $F^{\omega}_\beta\uparrow$ & $S_\alpha\uparrow$ & $E_\phi\uparrow$  
		& $M\downarrow$ & $F^{\omega}_\beta\uparrow$ & $S_\alpha\uparrow$ & $E_\phi\uparrow$
		& $M\downarrow$ & $F^{\omega}_\beta\uparrow$ & $S_\alpha\uparrow$ & $E_\phi\uparrow$\\
		\midrule
		
		\checkmark & \checkmark & 
		& \textcolor{red}{0.0405} & \textcolor{red}{0.8528}                          
		& \textcolor{red}{0.8838} & \textcolor{red}{0.9359}                 
		& 0.0208 & 0.8035
		& 0.8735 & 0.9276 
		& 0.0296 & 0.8515                      
		& 0.8909 & 0.9369\\
		
		\checkmark &           & \checkmark 
		
		& 0.0415 & 0.8481
		& 0.8819 & 0.9347
		& \textcolor{red}{0.0196} & \textcolor{red}{0.8174}                      
		& \textcolor{red}{0.8815} & \textcolor{red}{0.9371} 
		& \textcolor{red}{0.0283} & \textcolor{red}{0.8589}                  
		& \textcolor{red}{0.8959} & \textcolor{red}{0.9417}\\
		\bottomrule
	\end{tabular}
	
\end{table*}

\begin{table*}[!htbp]
	\centering
	\renewcommand{\arraystretch}{1.2}
	\setlength{\tabcolsep}{3pt}
	\footnotesize
	\caption{Ablation analysis of our MMF module.}
	\label{tab:edit_MMF}
	\begin{tabular}{ccc|cccc|cccc|cccc}
		\toprule
		\multicolumn{3}{c|}{\textbf{Structure settings}} 
		& \multicolumn{4}{c|}{\textbf{CAMO (250 images)}} 
		& \multicolumn{4}{c}{\textbf{COD10K (2026 images)}} 
		& \multicolumn{4}{c}{\textbf{NC4K (4121 images)}} \\
		
		\textbf{Baseline} & \textbf{MMF Add} & \textbf{MMF Cat} 
		& $M\downarrow$ & $F^{\omega}_\beta\uparrow$ & $S_\alpha\uparrow$ & $E_\phi\uparrow$  
		& $M\downarrow$ & $F^{\omega}_\beta\uparrow$ & $S_\alpha\uparrow$ & $E_\phi\uparrow$
		& $M\downarrow$ & $F^{\omega}_\beta\uparrow$ & $S_\alpha\uparrow$ & $E_\phi\uparrow$\\
		\midrule

		\checkmark & \checkmark & 
		& 0.0426 & 0.8442                       
		& 0.8790 & \textcolor{red}{0.9370}             
		& 0.0206 & 0.8089
		& 0.8768 & 0.9324 
		& 0.0295 & 0.8548                      
		& 0.8931 & 0.9396\\
		
		\checkmark &           & \checkmark

		& \textcolor{red}{0.0415} & \textcolor{red}{0.8481}
		& \textcolor{red}{0.8819} & 0.9347
		& \textcolor{red}{0.0196} & \textcolor{red}{0.8174}                      
		& \textcolor{red}{0.8815} & \textcolor{red}{0.9371} 
		& \textcolor{red}{0.0283} & \textcolor{red}{0.8589}                  
		& \textcolor{red}{0.8959} & \textcolor{red}{0.9417}\\
		\bottomrule
	\end{tabular}
	
\end{table*}
\begin{table*}[!htbp]
	\footnotesize
	\centering
	\setlength{\tabcolsep}{0.5pt}
	\renewcommand{\arraystretch}{1.2}
	
	\setlength{\tabcolsep}{2.8pt} % 控制列间距
	\caption{Params and Flops analysis of our DRRNet model and existing COD methods.}
	\label{tab:params}
	\begin{tabular}{c|cccccccccc}
		\hline
		& SINet\cite{ref11}& PFNet\cite{ref67} & MGL\cite{ref68}  & UGTR\cite{ref69} & UEDG\cite{ref58} & SDRNet\cite{ref20} & MVGNet\cite{ref61}  & FSPNet\cite{ref45}  & GLCONet-P\cite{ref33} & Ours \\
		\hline

		Params (M) & 48.95 & 46.50 & 63.60 & 48.87 & \textcolor{red}{17.93} & 90.54 & 86.62 & 273.79 & 91.35 & \textbf{89.11}\\
		
		FLOPs (G) & 38.75 & 53.22 & 553.94 & 1000.01 & \textcolor{red}{31.68} & 126.04 & 134.62 & 283.31 & 173.58 &\textbf{113.51}\\
		\bottomrule
	\end{tabular}
	
\end{table*}
\begin{figure*}[ht]
	\centering
	\includegraphics[width=0.8\textwidth]{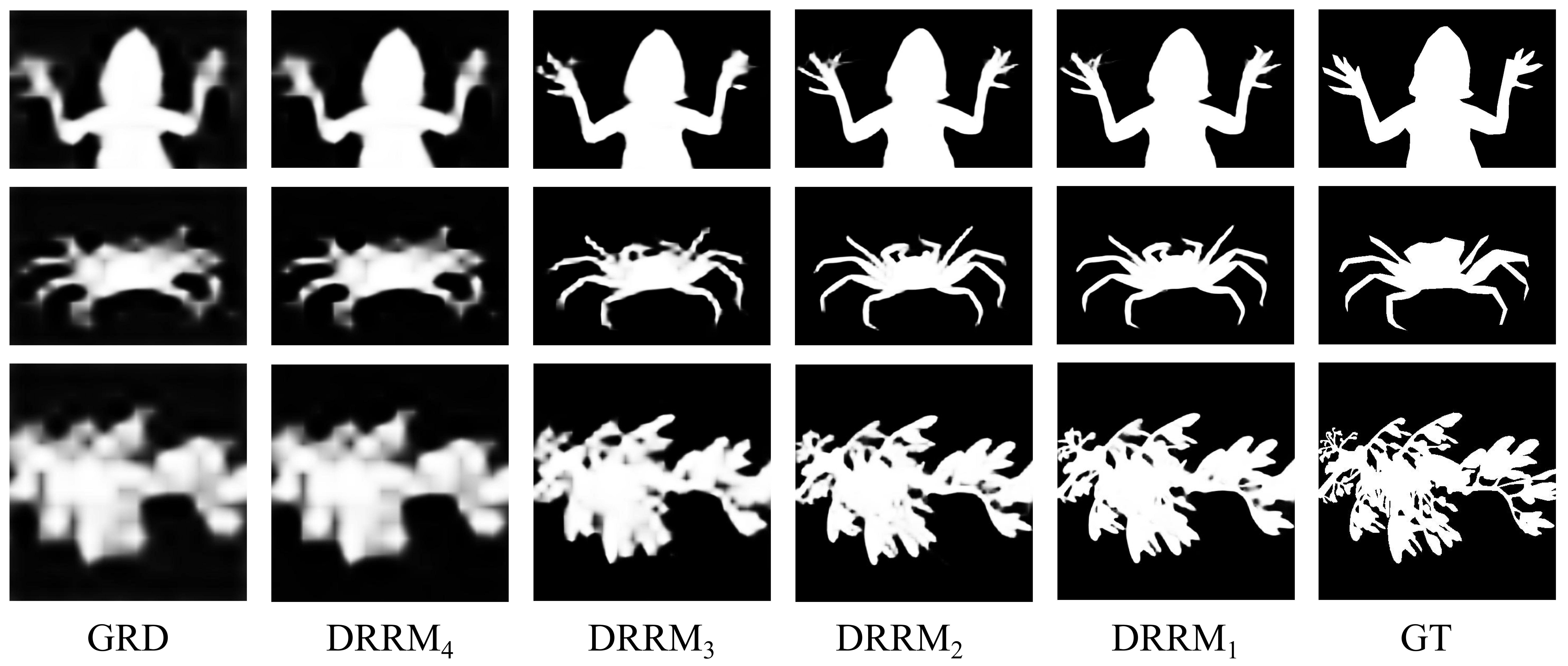}
	\caption{Visual results of the different prediction output of our model.}
	\label{fig:ablation_pre}
\end{figure*}

Ablation experiments on key components were conducted to validate their effectiveness and analyze their impact on performance, with results summarized in Table.~\ref{tab:ablation}. Notably, for the baseline model, all additional modules were removed, and basic convolutional blocks were used to replace original modules for prediction generation. Experimental results demonstrate that our designed feature extraction blocks (OCM, MDM), feature fusion block (MMF), and feature decoding blocks (GRD, DRRM) all significantly enhance detection performance. When combined to construct DRRNet, all evaluation metrics achieve substantial improvements.

\textbf{Effectiveness of Encoding Modules}. As shown in the 2nd and 3rd rows of Table.~\ref{tab:ablation}, compared with the baseline model, the encoding modules of OCM and MDM achieve significant performance improvements. According to Table.~\ref{tab:ablation}, adding OCM and MDM individually leads to average metric improvements of 1.67\% (MAE), 1.95\% ($F^{\omega}_{\beta}$), 3.16\% ($S_{\alpha}$), 3.34\% ($E_{\phi}$) and 1.3\% (MAE), 2.92\% ($F^{\omega}_{\beta}$), 4.01\% ($S_{\alpha}$), 3.58\% ($E_{\phi}$), respectively. Through detailed analysis, we attribute OCM’s improvement to its multi-scale global extraction of backbone features, which fully interacts and accumulates rich panoramic contextual semantic information to enhance detection accuracy. MDM, meanwhile, strengthens local information understanding via depthwise separable convolutions and dilated convolutions, significantly boosting feature representation capabilities for camouflaged object detection (COD). The 4th row of Table.~\ref{tab:ablation} further shows that combining OCM and MDM allows the model to learn panoramic and local information synergistically, further improving detection accuracy.

\textbf{Effectiveness of Feature Fusion Module}. For the BaseLine+OCM+MDM combination, feature fusion was initially performed via direct summation. After adding the MMF module, features undergo deep fusion through grouped frequency domain transformation and gated screening operations, leading to consistent improvements in all evaluation metrics.

\textbf{Effectiveness of Decoding Modules}. After learning rich features, a carefully designed decoder is essential for feature reconstruction. The GRD module generates coarse feature maps through multi-scale decoding of the deepest fused features. As shown in the 6th row of Table.~\ref{tab:ablation}, compared with using only convolutions, GRD achieves average metric improvements of 0.12\% (MAE), 1.2\% ($F^{\omega}_{\beta}$), 1.02\% ($S_{\alpha}$), 0.48\% ($E_{\phi}$). We analyze that without GRD-generated coarse prediction maps, the DRRM module lacks iterative optimization guidance. Directly using DRRM without the coarse decoding block only yields marginal improvements (0.23\% MAE, 1.4\% $F^{\omega}_{\beta}$, 1.45\% $S_{\alpha}$, 0.71\% $E_{\phi}$). This confirms that the joint use of GRD and DRRM is critical: with their collaborative operation, DRRNet achieves optimal performance across all metrics.

Visual predictions of different modules are shown in Fig.~\ref{fig:ablation_module}. It can be observed that by progressively integrating our designed modules, the model’s predictions gradually approach the ground truth (GT) results, validating the synergistic effectiveness of the proposed architecture.

\subsubsection{Fusion strategy}
In addition, we investigated the fusion strategies (element‐wise addition versus concatenation) employed by the OCM, MDM, and MMF modules, as summarized in Table.~\ref{tab:edit_ocm}, Table.~\ref{tab:edit_MDM} and Table.~\ref{tab:edit_MMF}. As shown in Table.~\ref{tab:edit_ocm} for the OCM module, concatenation is preferred for multi‐scale feature fusion because it better preserves the original multi‐scale information. In the MDM module (Table.~\ref{tab:edit_MDM}), both concatenation and addition exhibit their own merits. In the MMF module (Table.~\ref{tab:edit_MMF}), we observe that concatenating the global feature $g_i$ and the local feature $l_i$ during the initial fusion stage yields better performance than element-wise addition. We attribute this to the fact that concatenation more effectively preserves the intrinsic characteristics of both panoramic semantics and fine-grained local details, thereby enabling richer feature representations in the subsequent GFB module. Collectively, these results substantiate the efficiency and soundness of our proposed feature‐encoding modules for the camouflaged object detection task, owing to the complementary interplay among OCM, MDM, and MMF.

\subsubsection{Params and FLOPS}
In Table.~\ref{tab:params}, we compare the number of parameters (Params) and computational complexity (FLOPs) of the proposed DRRNet with those of existing mainstream methods. The results indicate that DRRNet achieves a favorable balance between performance and model complexity, employing only 89.11M parameters. Compared to large-scale architectures—which, despite their high representational capacity, are prone to overfitting and demand extensive computational resources—and lightweight models such as SINet—which offer low resource consumption but suffer from limited representational power—DRRNet demonstrates an effective trade-off between network scale and training efficiency. Moreover, DRRNet requires only 113.51G FLOPs, which is significantly lower than that of most competing approaches. This low computational overhead, combined with high accuracy, makes DRRNet particularly well-suited for deployment in resource-constrained environments (e.g., edge devices and mobile platforms), enabling faster and more efficient inference.

\subsubsection{Prediction Selection}
At the decoder stage, we designs a coarse decoder and a four - layer calibration decoder. Each layer of the decoder outputs a prediction image, but we only use the output of the last layer as the final result. To explore the role of our multi - layer decoder and prove the superiority of the decoder we designed, we designed additional experiments. In these experiments, the output of each layer of the decoder was taken as the final result for calculation. Some representative images are shown in Figure.~\ref{fig:ablation_pre}. It can be observed that with the progression of reverse optimization, the interference of the background on segmentation is significantly reduced. The predicted images gradually become clearer, the overall contours and edge information gradually become more distinct, and the false - positive regions are markedly decreased.

\subsection{Expanded Applications}

\begin{table*}[ht]
	\centering
	\caption{Quantitative results on four polyp segmentation datasets. The best result is shown in red. }
	\scriptsize
	\setlength{\tabcolsep}{0.5pt}
	\renewcommand{\arraystretch}{1.2}
	\label{tab:ablation_polyp}
	\setlength{\tabcolsep}{2.5pt} % 控制列间距
	\small
	
	\resizebox{\textwidth}{!}{
		
		\begin{tabular}{c|ccccc|ccccc|ccccc|ccccc}
			\hline
			
			\multirow{2}{*}{\textbf{Method}} 
			& \multicolumn{5}{c|}{\textbf{CVC-300 (60 images)}} 
			& \multicolumn{5}{c|}{\textbf{ETIS (196 images)}} 
			& \multicolumn{5}{c|}{\textbf{Kvasir (100 images)}} 
			& \multicolumn{5}{c}{\textbf{CVC-ColonDB (380 images)}} \\
			& $\text{MAE} \downarrow$ & $AF_m \uparrow$ & $WF_m \uparrow$ & $S_m \uparrow$ & $E_m \uparrow$ 
			& $\text{MAE} \downarrow$ & $AF_m \uparrow$ & $WF_m \uparrow$ & $S_m \uparrow$ & $E_m \uparrow$ 
			& $\text{MAE} \downarrow$ & $AF_m \uparrow$ & $WF_m \uparrow$ & $S_m \uparrow$ & $E_m \uparrow$ 
			& $\text{MAE} \downarrow$ & $AF_m \uparrow$ & $WF_m \uparrow$ & $S_m \uparrow$ & $E_m \uparrow$ \\
			\hline
			UNet~\cite{ref16}      & 0.028 & 0.701 & 0.674 & 0.839 & 0.872 
			& 0.040 & 0.399 & 0.362 & 0.688 & 0.635 
			& 0.058 & 0.835 & 0.789 & 0.879 & 0.902 
			& 0.062 & 0.555 & 0.490 & 0.702 & 0.748 \\
			UNet++~\cite{ref77}    & 0.026 & 0.702 & 0.677 & 0.819 & 0.852 
			& 0.045 & 0.465 & 0.388 & 0.677 & 0.689 
			& 0.050 & 0.843 & 0.801 & 0.861 & 0.902 
			& 0.061 & 0.557 & 0.462 & 0.689 & 0.753 \\
			SFA~\cite{ref78}       & 0.065 & 0.353 & 0.341 & 0.821 & 0.806 
			& 0.109 & 0.255 & 0.231 & 0.557 & 0.515 
			& 0.075 & 0.715 & 0.670 & 0.782 & 0.842 
			& 0.049 & 0.393 & 0.366 & 0.629 & 0.634 \\
			ACSNet~\cite{ref79}    & 0.127 & 0.781 & 0.825 & 0.922 & 0.916 
			& 0.059 & 0.536 & 0.530 & 0.750 & 0.774 
			& 0.032 & 0.900 & 0.882 & 0.920 & 0.944 
			& 0.039 & 0.720 & 0.682 & 0.829 & 0.860 \\
			GLCONet~\cite{ref33}   & 0.009 & 0.831 & 0.830 & 0.911 & 0.937 
			& 0.030 & 0.573 & 0.533 & 0.754 & 0.801 
			& 0.024 & 0.842 & 0.823 & 0.872 & 0.924 
			& 0.040 & 0.760 & 0.684 & 0.860 & 0.858 \\

			Ours & \textcolor{red}{0.007} & \textcolor{red}{0.906} & \textcolor{red}{0.859} & \textcolor{red}{0.931} & \textcolor{red}{0.951} 
			
			& \textcolor{red}{0.021} & \textcolor{red}{0.806} & \textcolor{red}{0.726} & \textcolor{red}{0.861} & \textcolor{red}{0.861}

			& \textcolor{red}{0.022} & \textcolor{red}{0.940} & \textcolor{red}{0.904} & \textcolor{red}{0.925} & \textcolor{red}{0.953}

			& \textcolor{red}{0.029} & \textcolor{red}{0.832} & \textcolor{red}{0.768} & \textcolor{red}{0.861} & \textcolor{red}{0.886} 
			\\
			\hline

		\end{tabular}
	}
	
\end{table*}

\begin{figure}[!htbp]
	\centering
	\includegraphics[width=1\linewidth]{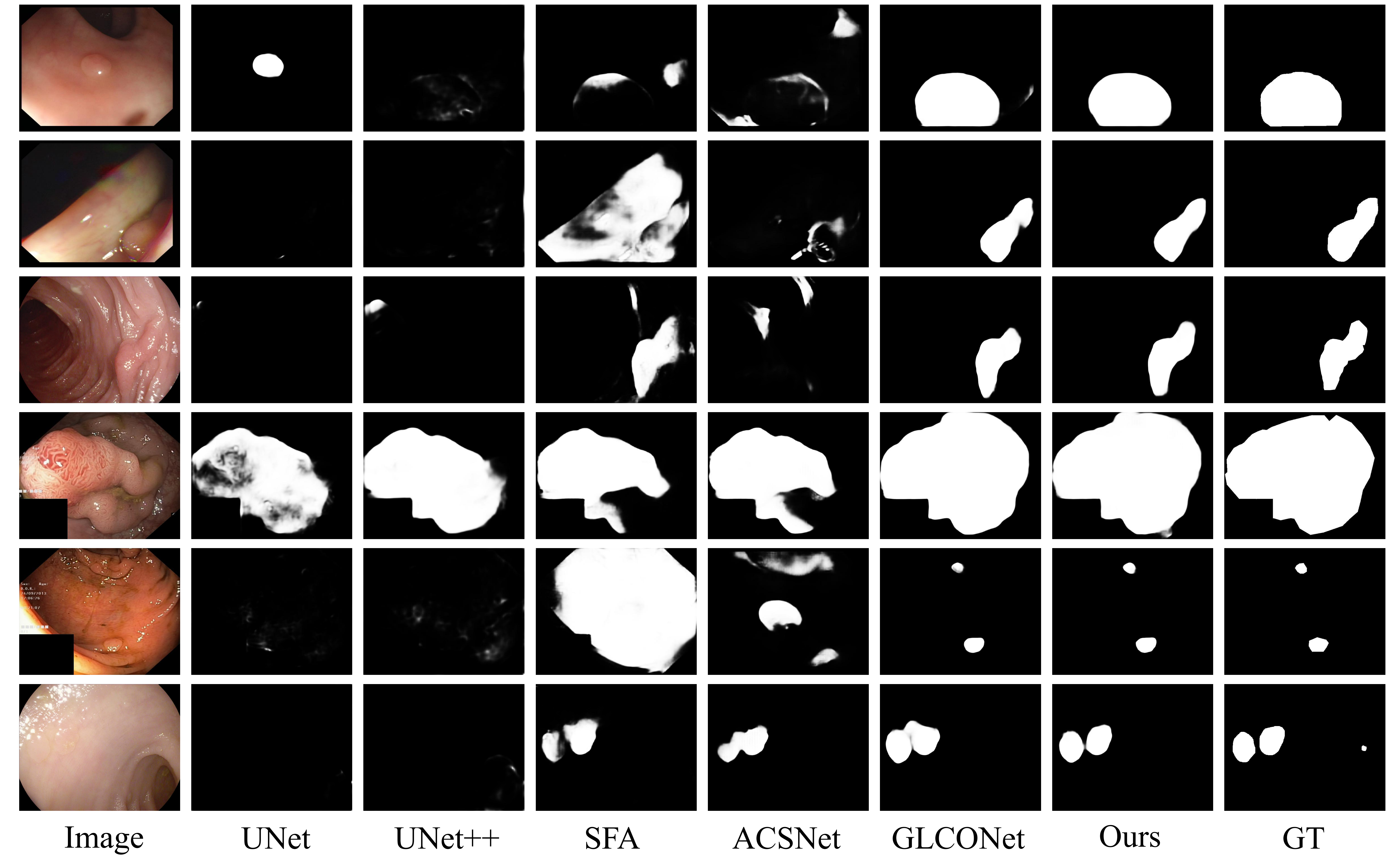}
	\caption{Qualitative visual comparisons of our DRRNet method and existing polyp segmentation methods.}
	\label{fig:ablation_polyp}
\end{figure}
Automatic polyp segmentation leverages computer vision algorithms to precisely analyze colonoscopic images, significantly reducing the likelihood of missed detections by medical professionals during colorectal polyp screening. This provides critical support for the early diagnosis and treatment planning of colorectal cancer. The technical challenges of this task are inherently similar to those in camouflaged object detection—specifically, the low contrast between polyp tissues and the surrounding intestinal background leads to weak feature discriminability. This necessitates the use of multi-modal feature fusion and contextual semantic modeling to achieve accurate segmentation. To further validate the effectiveness of our proposed DRRNet, we extend its application to the task of polyp segmentation.

\textbf{Experimental Details}. The parameter settings used in this work remain unchanged. We evaluate the performance of DRRNet on four public polyp segmentation datasets: CVC300\cite{ref72}, ETIS\cite{ref73}, Kvasir\cite{ref74}, and CVC-ColonDB\cite{ref75}. The DRRNet model is trained using the same 900 and 548 images from the ClinicDB\cite{ref76} and Kvasir datasets\cite{ref74}, respectively. To assess DRRNet's performance on the polyp segmentation task, we adopt the same five evaluation metrics used in the COD task: Mean Absolute Error (MAE), mean F-measure ($AF_m$), weighted F-measure ($WF_m$), S-measure($S_m$), and E-measure ($E_m$)\cite{ref49}.

\textbf{Results Comparison}. To verify the effectiveness of DRRNet for polyp segmentation, we compare it with five state-of-the-art (SOTA) methods: UNet\cite{ref16}, UNet++\cite{ref77}, SFA\cite{ref78}, ACSNet\cite{ref79}, and GLCONet\cite{ref33}. As shown in Table.~\ref{tab:ablation_polyp}, DRRNet significantly outperforms existing methods. In addition, visual comparison results are presented in Fig.~\ref{fig:ablation_polyp}, where DRRNet demonstrates precise segmentation of polyps within input images. These results collectively indicate that DRRNet exhibits strong generalization capability.

\subsection{Conclusion}
This paper addresses the issues of edge detail loss and background interference caused by the high similarity between targets and backgrounds in Camouflage Object Detection. We propose a four-stage framework named DRRNet, which achieves collaborative refinement of global semantics and local details through a ``panoramic perception - detail mining - cross-layer fusion - dual reverse calibration'' pipeline. Specifically, the Omni-Context Module captures scene-level camouflage patterns via a multi-branch attention mechanism, while the Micro-Detail Module enhances edge and texture features using cascaded dilated convolutions and channel attention. These modules undergo cross-layer feature fusion through the Macro-Micro Fusion Module, followed by spatial-frequency domain joint refinement via the Global Rough Decoder and Dual Reverse Refinement Module, forming a closed loop from feature extraction to result refinement. Comprehensive comparative experiments and ablation studies rigorously validate the effectiveness of the proposed DRRNet framework and its components. Furthermore, this paper extends DRRNet to the automatic polyp segmentation task. Experiments on datasets such as CVC300 and Kvasir demonstrate its remarkable generalization capability, effectively handling low-contrast segmentation issues between polyps and intestinal walls, which verifies the model's cross-task adaptability.

Although DRRNet demonstrates excellent performance across diverse scenarios, certain limitations persist in highly complex camouflage conditions—such as dynamic backgrounds and overlapping multi-target instances—where further improvements could be pursued. Future work plans to introduce dynamic feature selection mechanisms and explore the integration of multi-modal data (such as spectral and depth information) with the existing framework to further enhance the model's robustness to complex scenarios. Additionally, optimizing the lightweight design of the decoder to make it more suitable for real-time detection tasks is also an important research direction.

\section{Acknowledgments}
This research was supported in part by National Natural Science Foundation of China under grant No. 62061146001, 62372102, 62232004, 61972083, 62202096, 62072103, Jiangsu Provincial Key Research and Development Program under grant No. BE2022680, Fundamental Research Funds for the Central Universities (2242024k30022), Jiangsu Provincial Key Laboratory of Network and Information Security under grant No. BM2003201, and Collaborative Innovation Center of Novel Software Technology and Industrialization.

\newpage

\begin{IEEEbiography}[{\includegraphics[width=1in,height=1.25in,clip,keepaspectratio]{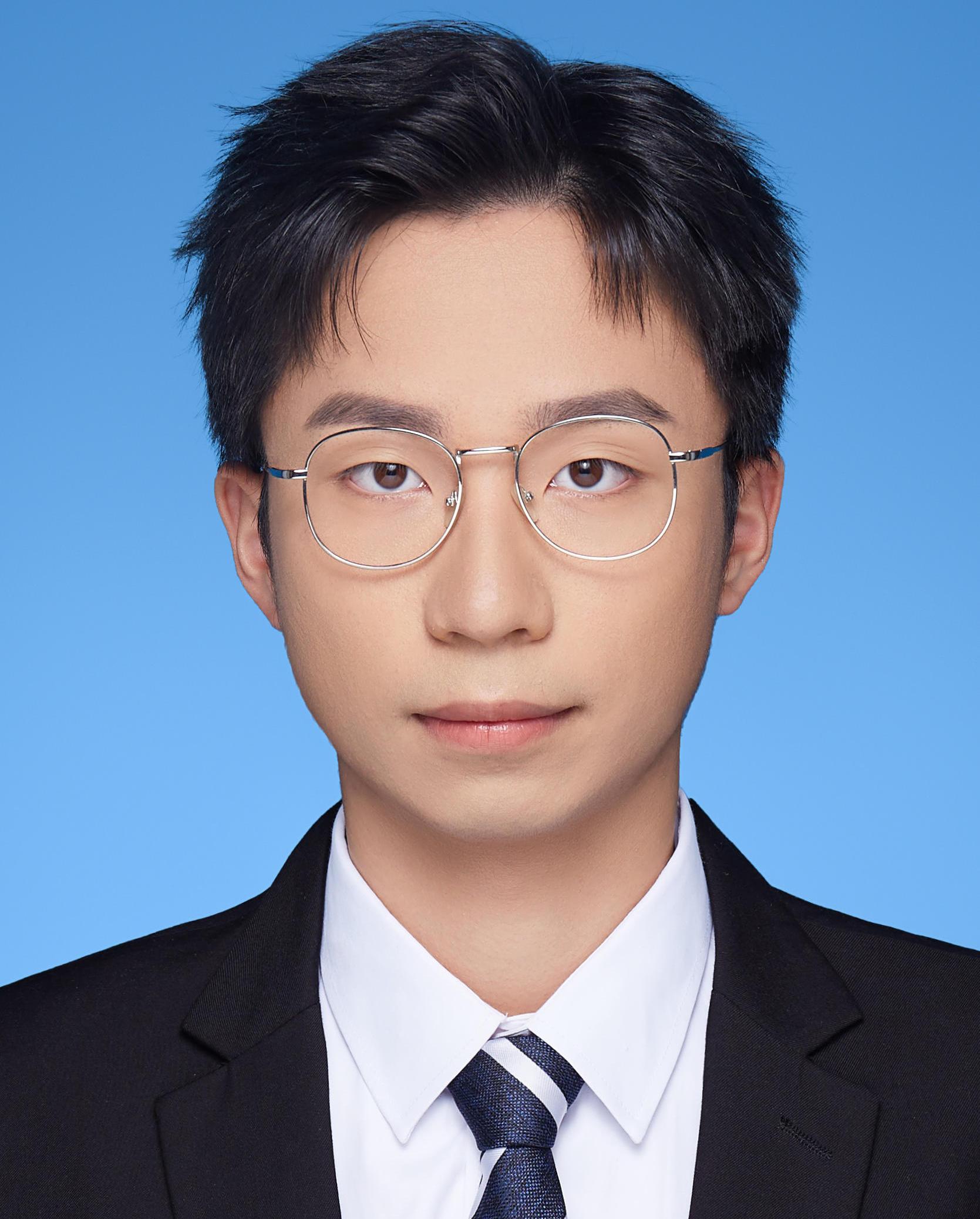}}]{Jianlin Sun}
	received the BS degree, in 2024 from the Hefei University of technology, Hefei, china. He is currently working toward the M.S. degree in School of Software with Southeast University, Nanjing, China. His research interests include deep learning, object detection and image segmentation.
\end{IEEEbiography}
\vspace{-33pt}

\begin{IEEEbiography}[{\includegraphics[width=1in,height=1.25in,clip,keepaspectratio]{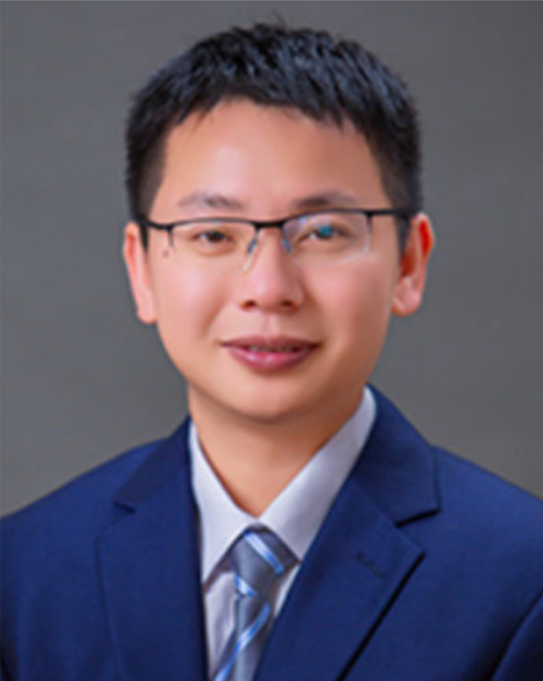}}]{Xiaolin Fang}
	(Member, IEEE) received the B.S. degree from Harbin Engineering University, Harbin, China, in 2007, and the M.S. and Ph.D. degrees from Harbin Institute of Technology, Harbin, in 2009 and 2014, respectively. He is currently an Associate Professor with the School of Computer Science and Engineering, Southeast University, Nanjing, China. His research interests include sensor networks, data processing, image processing, and scheduling.
\end{IEEEbiography}
\vspace{-33pt}
\begin{IEEEbiography}[{\includegraphics[width=1in,height=1.25in,clip,keepaspectratio]{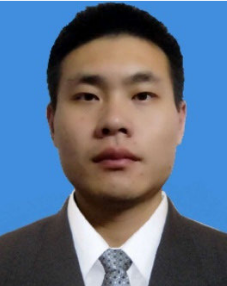}}]{Juwei Guan}
	received the M.S. degree in computer technology from Jiangxi University of Finance and Economics, Nanchang, China, in 2021. He is currently pursuing the Ph.D. degree with the School of Computer Science and Engineering, Southeast University, Nanjing, China. His current research interests include deep learning, image processing, domain generalization, and image segmentation.
\end{IEEEbiography}
\vspace{-33pt}
\begin{IEEEbiography}[{\includegraphics[width=1in,height=1.25in,clip,keepaspectratio]{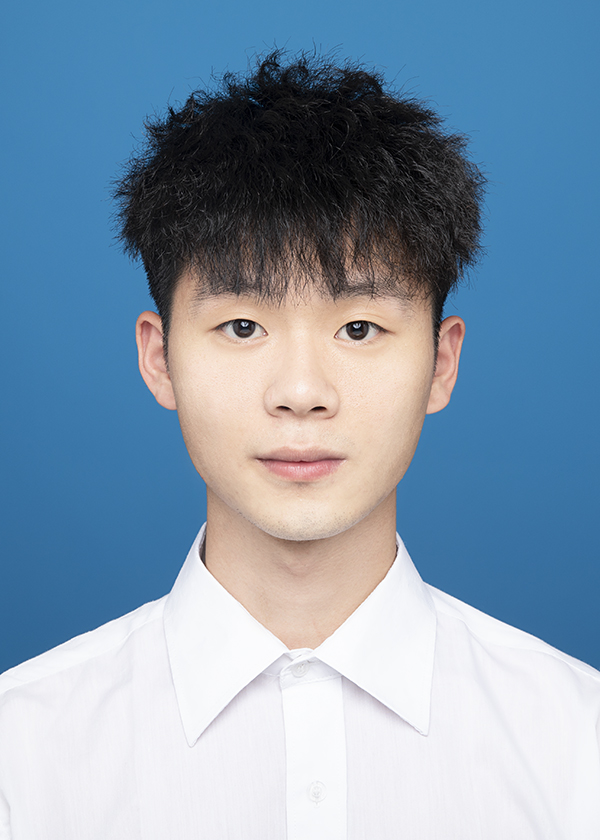}}]{Dongdong Gui}
	received the M.S. degree in Artificial Intelligence from the School of Software, Southeast University, Nanjing, China, in 2023. He is currently pursuing the Ph.D. degree with the School of Computer Science and Engineering, Southeast University, Nanjing, China. His current research interests include deep learning, image processing, and detection and segmentation in complex scenes.
\end{IEEEbiography}
\vspace{-33pt}
\begin{IEEEbiography}[{\includegraphics[width=1in,height=1.25in,clip,keepaspectratio]{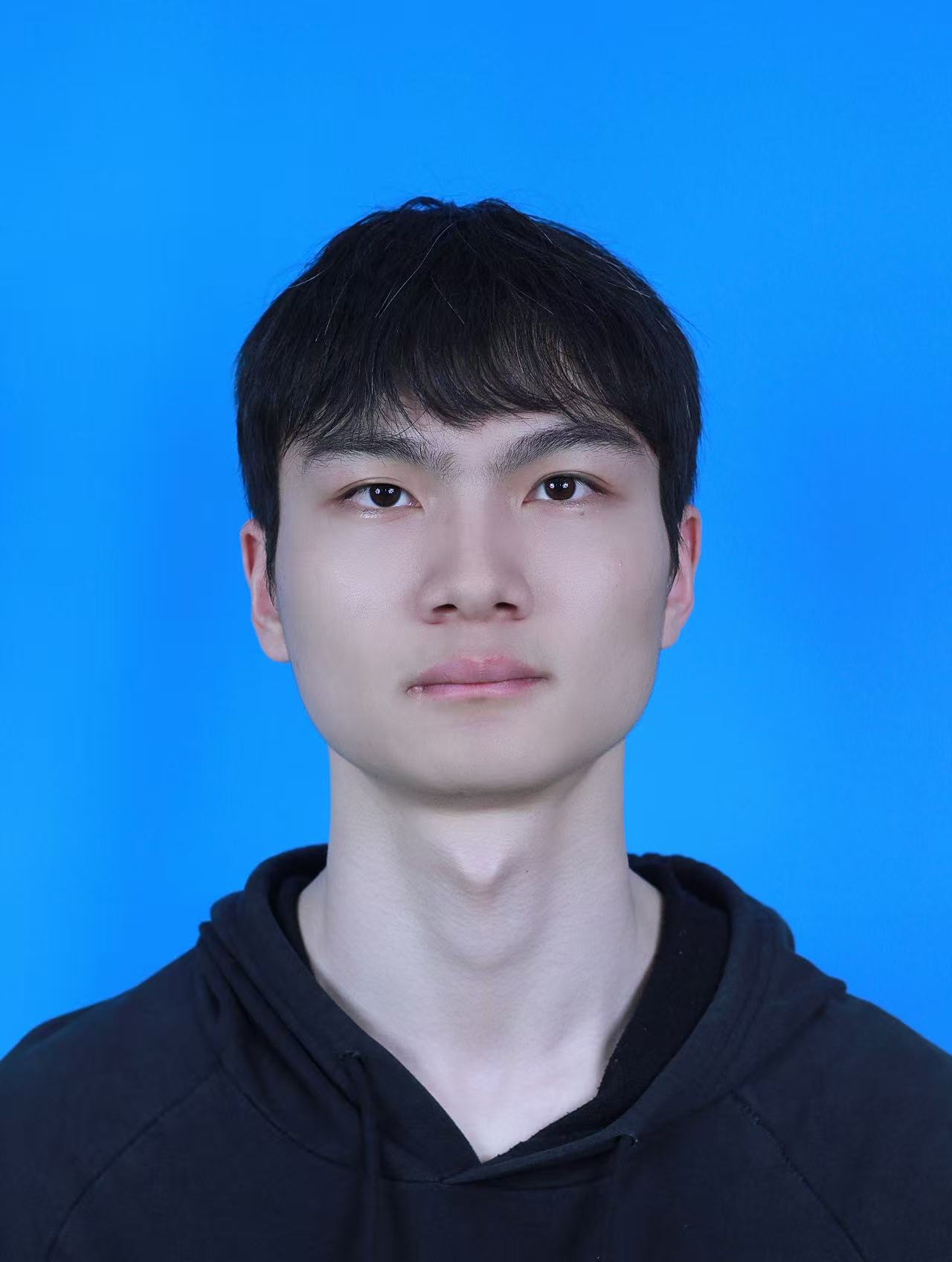}}]{Teqi Wang}
	received the M.S. degree in communication engineering from the University of Electronic Science and Technology of China, Chengdu, China, in 2023. He is currently pursuing the Ph.D. degree in the School of Computer Science and Engineering at Southeast University, Nanjing, China. His current research interests include model editing, image editing, and artificial intelligence.
\end{IEEEbiography}
\vspace{-33pt}
\begin{IEEEbiography}[{\includegraphics[width=1in,height=1.25in,clip,keepaspectratio]{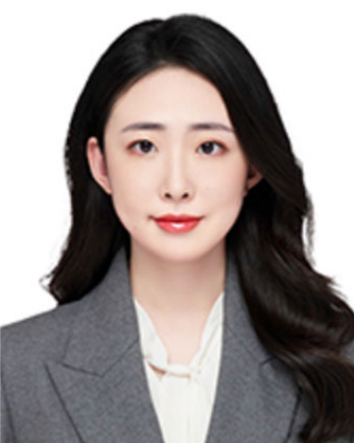}}]{Tongxin Zhu}
	(Member, IEEE) received the B.S. and Ph.D. degrees in computer science from Harbin Institute of Technology, China. She is currently a Lecturer with the School of Computer Science and Engineering, Southeast University, Nanjing, China. Her research interests include sensor networks, the Internet of Things, and mobile edge computing.
\end{IEEEbiography}

\end{document}